\documentclass{article}

\usepackage[preprint]{neurips_2025}

\usepackage[utf8]{inputenc} 
\usepackage[T1]{fontenc}    
\usepackage[pagebackref=true,breaklinks=true,colorlinks,bookmarks=false,citecolor=blue,linkcolor=blue]{hyperref}
\usepackage{url}            
\usepackage{booktabs}       
\usepackage{amsfonts}       
\usepackage{nicefrac}       
\usepackage{microtype}      
\usepackage{xcolor}         
\usepackage{multicol}
\usepackage{multirow}
\usepackage{booktabs}
\usepackage{caption}
\usepackage{subcaption}
\usepackage{tabularx}
\usepackage{newfloat}
\usepackage{listings}
\usepackage{tcolorbox}
\usepackage{algorithm}
\usepackage{algorithmic}
\usepackage{colortbl}
\usepackage{lscape}
\usepackage{graphicx}
\usepackage{pifont}
\usepackage{subcaption} 
\usepackage{wrapfig}
\usepackage{floatflt}
\usepackage{tcolorbox}
\usepackage{marvosym}
\usepackage{afterpage}
\usepackage{makecell}
\usepackage{subcaption}

\newcommand{\tm}{\textit{Tom \& Jerry}\xspace}

\usepackage{dsfont}
\usepackage{mathtools,thmtools}
\usepackage{xfrac}
\usepackage{multirow}
\usepackage{amsthm,amsmath,amssymb}
\usepackage{color, colortbl}
\usepackage{subcaption}
\usepackage{pifont}
\usepackage{bm}
\usepackage{xspace} 
\usepackage{wrapfig}
\usepackage{enumitem}
\usepackage{enumitem}
\usepackage[normalem]{ulem}
\useunder{\uline}{\ul}{}

\makeatletter
\DeclareRobustCommand\onedot{\futurelet\@let@token\@onedot}
\def\@onedot{\ifx\@let@token.\else.\null\fi\xspace}
\def\eg{\emph{e.g}\onedot} 
\def\ie{\emph{i.e}\onedot} 
 
\def\etc{\emph{etc}\onedot}

\makeatother

\definecolor{darkorange}{rgb}{1.0, 0.55, 0.0}

\definecolor{Gray}{gray}{0.9}
\definecolor{LightBlue}{RGB}{236,244,249}

\definecolor{myblue}{RGB}{65,115,176}
\definecolor{mygreen}{RGB}{126,171,86}
\definecolor{myyellow}{RGB}{249,187,0}

\newcommand{\CC}[1]{\cellcolor{LightBlue}}
\newcommand{\RC}[1]{\rowcolor{LightBlue}}

\newcommand{\cmark}{\color{green}\ding{51}}%
\newcommand{\xmark}{\color{red}\ding{55}}%

\definecolor{citecolor}{rgb}{0,0.443,0.737} 
\definecolor{linkcolor}{rgb}{0.956,0.298,0.235} 
\hypersetup{
    colorlinks=true,%
    citecolor=citecolor,%
    filecolor=citecolor,%
    linkcolor=linkcolor,%
    urlcolor=magenta
}

\usepackage{setspace}

\title{AudioStory: Generating Long-Form Narrative Audio with Large Language Models}

\author{
\vspace{2mm}
        Yuxin Guo\textsuperscript{1,2,3}, 
        ~Teng Wang\textsuperscript{2$\dagger*$},
        ~Yuying Ge\textsuperscript{2},
        ~Shijie Ma\textsuperscript{1,2,3},
        ~Yixiao Ge\textsuperscript{2},
        ~Wei Zou\textsuperscript{1,3$*$},
        ~Ying Shan\textsuperscript{2} \\ 
\vspace{2mm}
        $^1$School of Artificial Intelligence, University of Chinese Academy of Sciences \\
	$^2$ARC Lab, Tencent PCG\qquad $^3$MAIS, Institute of Automation, CAS, Beijing \\
\vspace{2mm}
        $^\dagger$ Project Lead \ \ \ $^*$ Corresponding Authors \\
    \url{https://github.com/TencentARC/AudioStory}
}

\begin{document}

\maketitle

\begin{abstract}

Recent advances in text-to-audio (TTA) generation excel at synthesizing short audio clips but struggle with long-form narrative audio, which requires temporal coherence and compositional reasoning. To fill this gap, we propose AudioStory, a unified framework that integrates large language models (LLMs) with TTA systems to generate structured, long-form audio narratives. AudioStory possesses strong instruction-following reasoning generation capabilities. It employs LLMs to decompose complex narrative queries into temporally ordered sub-tasks with contextual cues, enabling coherent scene transitions and emotional tone consistency. AudioStory has two appealing features: 
(1) Decoupled bridging mechanism: AudioStory disentangles LLM-diffuser collaboration into two specialized components, \ie, a bridging query for intra-event semantic alignment and a residual query for inter-event coherence preservation.
(2) End-to-end training: By unifying instruction comprehension and audio generation within a single end-to-end framework, AudioStory eliminates the need for modular training pipelines while enhancing synergy between components. 
Furthermore, we establish a benchmark AudioStory-10K, encompassing diverse domains such as animated soundscapes and natural sound narratives.
Extensive experiments show the superiority of AudioStory on both single-audio generation and narrative audio generation, surpassing prior TTA baselines in both instruction-following ability and audio fidelity. 

\end{abstract}

\section{Introduction}

Audio content plays a pivotal role in modern media, from immersive storytelling and podcasts to interactive entertainment and educational applications. Recent advancements in text-to-audio (TTA) generation, exemplified by models such as TangoFlux~\cite{hung2024tangoflux}, AudioLDM~\cite{liu2024audioldm2learningholistic}, and Stable Audio~\cite{evans2024stableaudioopen}, have demonstrated remarkable capabilities in synthesizing high-quality, short-form audio clips from textual descriptions. 
However, a critical gap remains in generating long-form narrative audio, \ie, coherent, structured sequences of audio instances that unfold over extended durations, such as audiobooks, podcasts, or dynamic soundscapes for games.  

Long-form narrative audio generation introduces unique challenges that extend beyond single-prompt synthesis. First, it requires temporal coherence: maintaining consistency in themes, sound effects, and emotional tone across the whole audio. Second, it demands narrative reasoning to decompose a complex instruction into logically ordered sub-events, characters, or environmental interactions. For instance, a prompt like ``A suspenseful chase through a rainstorm: footsteps splash, thunder roars, a car skids, and a door slams shut'' necessitates not only generating individual sounds but also orchestrating their timing, intensity, and interplay to build tension. Existing TTA models, while proficient at capturing isolated events, often struggle with such compositional and temporal reasoning, leading to fragmented or inconsistent outputs.

\begin{figure}
    \centering
    \includegraphics[width=1.0\linewidth]{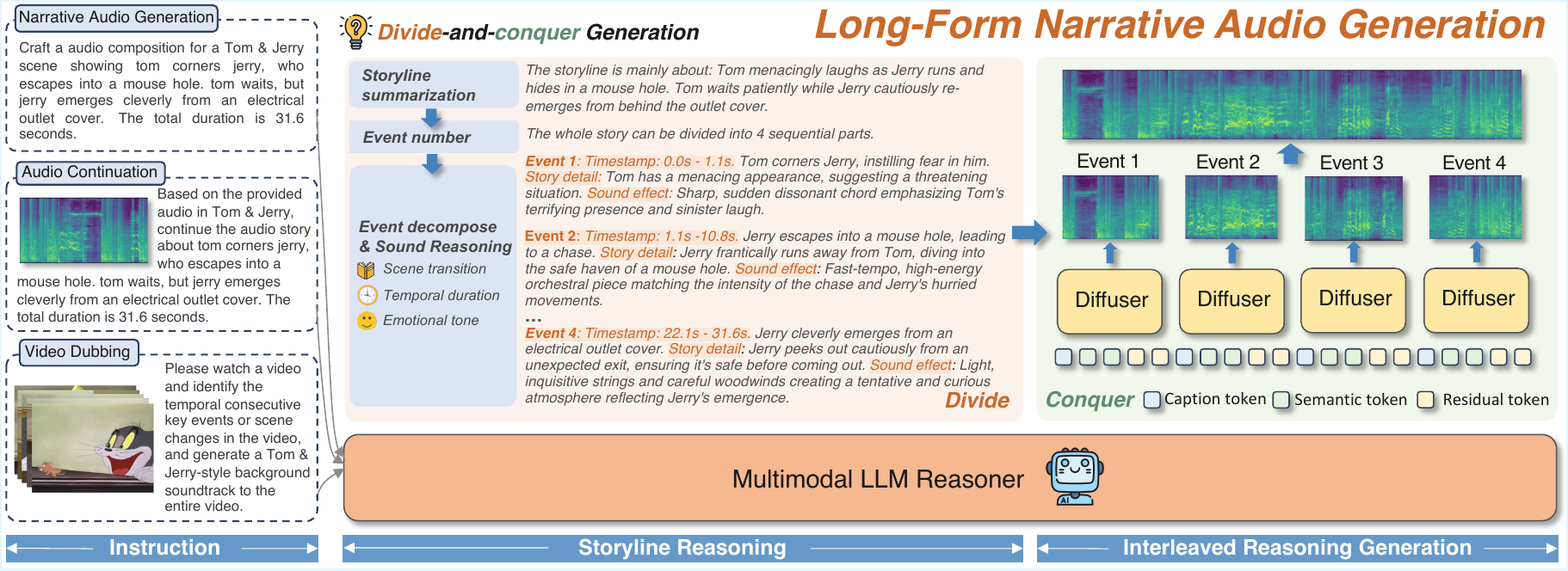}
    \caption{AudioStory effectively follows multimodal instructions, decomposing them into a sequence of coherent audio segments, capturing scene transitions, emotional tone, and segment timestamps. Unlike prior T5-based diffusion models, which struggle with complex queries, AudioStory empowers LLMs with high-level planning ability for instruction-followed and consistent long audio generation.}
    \label{fig:intro}
    \vspace{-10pt}
\end{figure}

To address these challenges, we propose AudioStory, a novel multi-step framework for generating long-form narrative audio by integrating the reasoning capabilities of LLMs with audio generation. As shown in Fig.~\ref{fig:intro}, we propose \textit{interleaved reasoning generation} following a divide-and-conquer manner: reasoning for general narrative plans, decomposing plans into sequential generation actions, and generating interleaved audio events step-by-step. Specifically, AudioStory employs LLMs to decompose a narrative query (in language or multimodality) into a structured sequence of audio-generative sub-tasks, each accompanied by contextual cues such as temporal offsets, emotional tone, and character interactions. These reasoning chains are then synthesized into audio events using a diffusion backbone, with explicit mechanisms to ensure style consistency, smooth transitions and temporal alignment. We streamline the narrative planning via LLMs and audio synthesis via diffusion models into an end-to-end framework, enabling the generation of rich, multi-scene audio stories that adhere to user intent while preserving coherence over time.

AudioStory introduces several technical innovations: First, unlike prior approaches~\cite{wu2024nextgpt, lai2024spider} that bridge LLMs with audio diffusers through predefined textual spaces (\emph{e.g.}, T5~\cite{raffel2023exploringlimitstransferlearning}), we propose a decoupled bridging space consisting of two distinct tokens: (1) \textit{semantic tokens}, which encode text-oriented audio semantics, and (2) \textit{residual tokens}, which capture nuanced acoustic cues and cross-event correlations. This design effectively improves both audio fidelity and temporal consistency during generation.
Second, unlike zero-shot integration of LLMs and diffusers, our framework supports end-to-end progressive training, enabling joint optimization of instruction understanding and audio synthesis. This synergistic training paradigm enhances both audio understanding and generation performance.
Third, we introduce the first narrative audio generation benchmark, providing a comprehensive evaluation framework for assessing audio generation quality and consistency.

The contributions of the paper are as follows:
\begin{itemize}[leftmargin=*]
\item  We introduce AudioStory for narrative audio generation, which integrates LLM-based reasoning and iterative diffusion-based generation in a unified framework, with strong multimodal instruction-following and audio generation abilities.

\item We propose decoupled bridging tokens for LLM-diffuser collaboration, using semantic tokens (text-oriented audio semantics) and residual tokens (nuanced acoustic cues) to improve audio fidelity and temporal consistency.

\item We introduce a synergistic training paradigm, facilitating collaboration and complementarity between LLM and diffusion models. Unlike zero-shot LLM-diffusion integration, our framework enables end-to-end joint training, enhancing both multimodal understanding and generation. 

\item Experiments show AudoStory significantly surpasses prior diffusion-based and MLLM-based models by a large margin in narrative audio generation. We also uncover some important findings across multiple aspects, including reasoning formulation, bridging mechanism and training recipes.
\end{itemize}

\section{Related Works}

\noindent \textbf{Text-to-audio generation (TTA).}
Recent advances in latent diffusion and flow-matching frameworks have significantly advanced text-to-audio generation. \textit{Diffusion/flow-based approaches}, exemplified by Make-An-Audio~\cite{huang2023makeanaudiotexttoaudiogenerationpromptenhanced} and AudioLDM~\cite{liu2023audioldmtexttoaudiogenerationlatent,liu2024audioldm2learningholistic}, synthesize audio through iterative denoising of text-conditioned latent representations. Extensions like Tango~\cite{majumder2024tango2aligningdiffusionbased, ghosal2023texttoaudiogenerationusinginstructiontuned}, Audio Flamingo~\cite{kong2024audioflamingonovelaudio}, GenAu~\cite{haji2024taming}, Fugatto~\cite{valle2025fugatto} further enhance design spaces of latent space, data quality and cross-modal alignments. Recently, Stable Audio series~\cite{evans2024stableaudioopen} employs hierarchical latent diffusion trained on large-scale datasets for high-fidelity output. Beyond diffusion-based priors, flow-matching techniques optimize probability density transport for audio synthesis. VoiceBox~\cite{le2023voiceboxtextguidedmultilingualuniversal} enables zero-shot style transfer via continuous normalizing flows, while AudioBox~\cite{vyas2023audioboxunifiedaudiogeneration} and FlashAudio~\cite{liu2024flashaudiorectifiedflowsfast} prioritize computational efficiency through rectified flow architectures. TangoFlux~\cite{hung2024tangoflux} introduces CLAP-ranked preference optimization to iteratively generates and optimizes preference data to enhance text-audio alignment. 
Existing methods align text and audio semantically but primarily target descriptive queries, limiting interactive control and adaptability to evolving instructions.
They are also confined to short audio domains. These limitations demand TTA models to handle complex instructions over long durations.

\noindent \textbf{Any-to-any multimodal LLMs.}
Within the rapidly evolving field of multimodal learning, \emph{any-to-any} generation across vision, language, and audio modalities represents a significant frontier~\cite{jin2023unified,tang2023codi2,wu2024nextgpt,zhan2024anygpt,lai2024spider,ge2023seed-llama,wu2024vila-u,xie2024show-o,team2024chameleon, lin2025toklip, ma2025genhancer, guo2025aligned}. This paradigm aims for models capable of accepting arbitrary input modalities and generating outputs in any desired modality. Pioneering efforts include CoDi~\cite{tang2023codi} and CoDi-2~\cite{tang2023codi2}, which leveraged composable diffusion for diverse modality handling. Spider~\cite{lai2024spider} further extended these capabilities by enabling the generation of multiple modalities in a single response. NExT-GPT~\cite{wu2024nextgpt} demonstrated the efficacy of lightweight alignment for adapting LLMs to multimodal tasks, while AnyGPT~\cite{zhan2024anygpt} showcased the potential of discrete sequence modeling for unified multimodal processing. Unified-IO2~\cite{lu2023unifiedio2} highlighted the impact of scale and unified architectures in achieving state-of-the-art performance across a broad spectrum of modalities and tasks. Despite these advancements, current methods exhibit limitations in long-context generation with complex instructions: First, they primarily focus on speech generation and simple caption-to-music or caption-to-sound tasks, struggling to comprehend general and intricate human instructions beyond basic caption; Second, their audio generation is typically limited to single, short segments, hindering the generation of longer audio sequences.

\noindent \textbf{Compositional audio generation.} Agentic workflows employ multiple off-the-shelf expert tools and a controller for compositional audio synthesis.  
Works like WavJourney~\cite{liu2025wavjourney} and MM-StoryAgent~\cite{xu2025mm} decomposed audio generation into a text-centric interface and employs separate text-to-speech, audio, and music decoders for audio creation and storytelling. 
While these agents could generate combinations of audio components, their zero-shot nature suffers from suboptimal planning and limited adaption of nuanced acoustic cues, degrading instruction-following ability and overall audio quality.  
Instead, we target on end-to-end training to integrate LLM-based chain-like reasoning and flux decoders for long-term, consistent audio generation.

\section{Narrative Audio Generation}
\label{sec:audiostory-10k}

\noindent \textbf{Problem definition.}
Narrative audio generation aims to generate long-form, structured and temporally coherent audio sequences $A={\{A_m\}}_{m=1}^{M}$, given multimodal instruction $x_{\rm ins}$ (\eg, language, audio or vision), where $M$ is the number of audio segments. The task shares a similar formulation with the text-to-audio generation, but is far more challenging due to two distinct capabilities: (1) Temporal coherence, \ie, maintaining consistency in themes, sound effects, and emotional tone across extended durations; (2) Compositional reasoning. \ie, decomposing high-level narrative instructions into logically ordered events (\eg, ``footsteps splash, then thunder roars'') with precise timing and contextual interactions. Existing TTA systems, while effective for short clips, lack explicit mechanisms to model cross-segment dependencies or align audio events with evolving narrative structures, limiting their applicability to real-world scenarios.

\noindent \textbf{The AudioStory-10k benchmark.}
Given the lack of quantitative evaluation, we establish the AudioStory-10k benchmark for the narrative audio generation task. AudioStory-10k comprises 10k annotated audios paired with narrative prompts. We collect videos from two primary sources:

\begin{itemize}[leftmargin=*]
    \item \textbf{Natural sounds}: We carefully select 4,723 audio instances from UnAV-100~\cite{geng2023dense}, covering a broad spectrum of real-world environmental recordings (\eg, rainstorms, animal calls, rustling leaves) and human activities (\eg, footsteps, door slams, and conversations). This collection ensures sufficient coverage of everyday acoustic events and ambient soundscapes.
    \item \textbf{Animated sounds}: We curate 5,332 audio clips from 157 episodes of \textit{Tom \& Jerry}, capturing stylized background music (\eg, orchestral pieces, string sections) and sound effects (\eg, slapstick actions, cartoonish collisions and rapid movements). These animated sounds feature stylized and expressive audio content, distinct from natural sound recordings.
\end{itemize}

The annotation pipeline involves three stages. First, we filter the videos with sequential audio events, ensuring the storyline of the audio is visually-grounded for meaningful activities\footnote{For \tm, where episodes typically consist of numerous discontinuous shots with fast transitions, we employ PySceneDetect~\cite{Castellano2018} to detect preliminary shot boundaries. These boundaries are further refined by thresholding the similarity between frame-level DINOv2~\cite{oquab2023dinov2} features, which retains only high-quality, temporally consistent shots as individual video instances. For UnAV-100, we keep the videos longer than 30s.}.
Then, we parse the video into several key audio events by Gemini-2.5-Pro~\cite{team2023gemini}, each of which is labeled with its timestamps, audio caption and visual captions. Next, given these text-based timestamped captions, we prompt GPT-4o~\cite{openai2025gpt4o} to generate diverse instructions and chain-like reasoning steps. 

To be specific, we design diverse format of multimodal instructions, including text-only instructions for narrative audio generation, audio-text ones for audio continuation and video-text ones for video dubbing (shown in Fig.~\ref{fig:intro}). For a flexible control of duration and semantic elements of generated audios, we make the intermediate reasoning encompass at least the following steps: \textit{storyline summarization} for global summarization of general story, \textit{event decomposition} for inferring the number of audio events, \textit{sound reasoning} for predicting timestamp and key elements (\eg, emotional tone, scene transition) of each event. All detailed prompts and processing steps are in Appendix~\ref{appendix:benchmark-dataset}. 

\noindent \textbf{Evaluation metrics.}
The AudioStory-10k dataset includes 5.3k samples of natural sounds and 4.7k samples of cartoon audios. We randomly divided the dataset into 85\% for training and 15\% for testing. We propose a comprehensive evaluation spanning three aspects: \textit{instruction-following}, \textit{consistency}, and \textit{generation quality}. (1) {Instruction-following ability} is quantified through multimodal alignment between instructions and generated audio
(\textit{Instruct}), CLAP score for audio-caption similarity, and \textit{reasoning text quality} for logical decomposition and event planning. (2) {Consistency metrics} evaluates internal \textit{consistency} (timbre uniformity, entity persistence) and temporal \textit{coherence} (acoustic transitions, emotional flow). (3) {Generation quality metrics} employs FD and FAD~\cite{kilgour2018fr} against ground-truth audio. Except from CLAP, FD, FAD-based metrics, we employ Gemini-2.0-flash as the evaluator with a score range of 0-5. More details could be found in the Appendix~\ref{appendix:benchmark-eval}.

\begin{figure}[!t]
    \centering
    \includegraphics[width=1.0\linewidth]{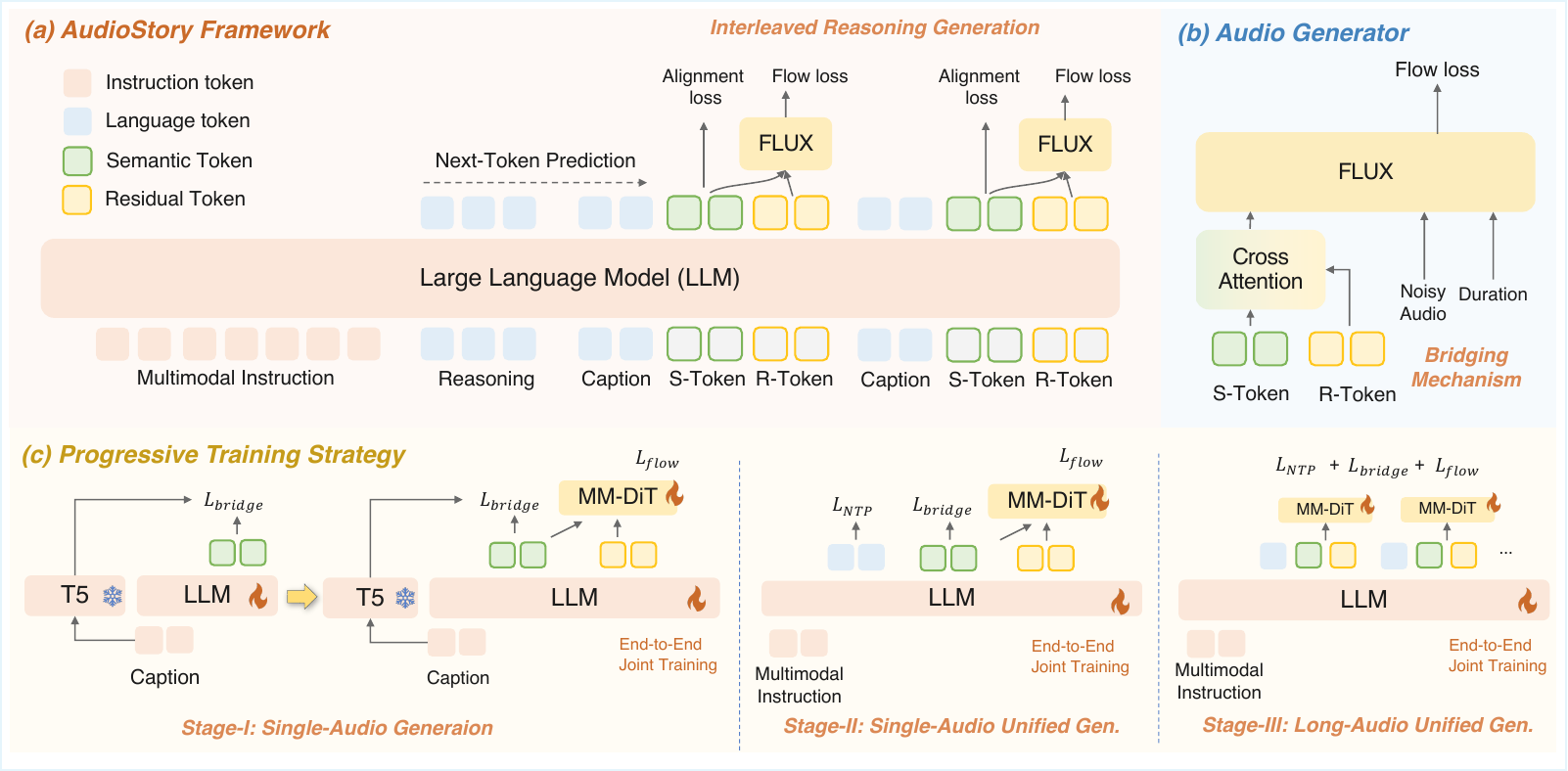}
    \caption{Overview of AudioStory, with three core components: (a) A unified framewrok: The reasoning-capable LLM processes the instruction input, decomposes the long audio into structured generation sub-tasks, and sequentially generates a caption, semantic tokens, and residual tokens for each audio clip. (b) Audio Generator: After fusing semantic and residual tokens, they are combined with the duration information as conditioning inputs to the DiT, which then generates each audio clip. (c) Training strategy: Training is conducted in three stages to progressively enhance generation fidelity, semantic understanding, and global coherence.
    }
    \label{fig:method}
\end{figure}

\section{AudioStory}

\vspace{-5pt}
\noindent \textbf{Overview.}
To achieve instruction-followed audio generation, the ability to understand the input instruction and reason about relevant audio sub-events is essential. To this end,  AudioStory adopts a unified understanding-generation framework (Fig.~\ref{fig:method}). Specifically, given multimodal instructions, an LLM analyzes and decomposes it into structured audio sub-events with context. Based on the inferred sub-events, the LLM first performs interleaved reasoning generation (Sec.~\ref{subsec:inter-reason-gen}), sequentially producing captions and bridging tokens between the LLM and the audio generator~(Sec.~\ref{subsec:bridge}). Through progressive end-to-end training, AudioStory ultimately achieves both strong instruction comprehension and high-quality audio generation (Sec.~\ref{subsec:train}).

\subsection{Interleaved Reasoning Generation} \label{subsec:inter-reason-gen}
Directly generating long-form narrative audio that aligns with complex instructions is challenging. We take the spirit of ``divide-and-conquer'' and propose decoupling the input instruction into chronological short audio clips, which are then combined to form the complete long-form narrative audio.

\noindent \textbf{Single-audio clip generation.}
The ability to generate individual audio clips from captions is a foundational step toward producing sequential audio events. For audio clip generation, the LLM generates bridge tokens from a given caption, which serve as conditions for the DiT. While this method works well for short audio generation based on simple captions, it becomes insufficient for complex instructions involving multiple events, temporal relationships, or narrative structures.

\noindent \textbf{Interleaved reasoning generation for long-audio generation.}
We propose to decouple a complex, long-form audio into multiple audio segments for segment-by-segment generation. This divide-and-conquer process consists of two components: (1) \emph{Storyline reasoning}: LLMs reason through the entire instruction, inferring the number of audio events. Furthermore,  LLMs analyze the start and end timestamps of each event, as well as the event description and corresponding audio content that should be included. 
(2) \emph{Interleaved generation}: For each event, the LLM infers the caption, duration, and corresponding bridge queries (semantic tokens and residual tokens, as described in Sec.~\ref{subsec:bridge}), enabling interleaved generation. These queries, along with duration information, are then provided as conditional inputs to the DiT-based audio generator. By accurately predicting durations and utilizing semantically rich bridging tokens, the model ensures both coherent audio semantics within each event and consistency across events. The training data is structured as: 
\begin{equation}
\begin{aligned}
    &\texttt{[BOS]} \texttt{[BOT]} {\color{myblue}\{\text{\#event}\}\{\text{storyline reasoning tokens}\}} \texttt{[EOT]}
    \texttt{[BOG]} {\color{myblue}\{\text{caption}\}} {\color{myblue}\{\text{duration}\}}\\ &{\color{mygreen}\mathbf{T}_\text{semantic}} {\color{myyellow}\mathbf{T}_\text{residual}} \texttt{[EOG]} \cdots
    \texttt{[BOG]} {\color{myblue}\{\text{caption}\}} {\color{myblue}\{\text{duration}\}} {\color{mygreen}\mathbf{T}_\text{semantic}} {\color{myyellow}\mathbf{T}_\text{residual}} \texttt{[EOG]}\texttt{[EOS]}.
    \label{eq:template}
\end{aligned}
\end{equation}
The textual tokens in the entire reasoning process is supervised by the next token prediction loss:
\begin{equation}
    \mathcal{L}_\text{reason}=\mathcal{L}_\text{text}^\text{\#event}+\mathcal{L}_\text{text}^\text{content}+\mathcal{L}_\text{text}^\text{caption},\quad
    \text{where}\quad\mathcal{L}_\text{text}=\prod_{i=1}^L p(\boldsymbol{x}_i|\mathbf{X}_{<i},\mathbf{X}_{p,<i}).
\end{equation}

\subsection{Decoupled Bridging Mechanism}
\label{subsec:bridge}

Once the LLM is capable of effective reasoning, establishing a seamless bridge between the LLM and the DiT becomes crucial. However, text \emph{alone} might not be the optimal bridge. Although it carries rich semantics, it fails to capture diverse low-level details of the audio modality, \eg, timbre, rhythm, and ambience. Consequently, we propose decoupled bridges queries, which could be divided into semantic $\mathbf{T}_\text{semantic}$ and residual tokens $\mathbf{T}_\text{residual}$. The semantic tokens represent the audio's high-level semantics, while the residual tokens carry low-level audio details. They complement each other, enabling the disentanglement of audio information.
In practice, after producing the caption for each audio event, the LLM collectively generates semantic and residual tokens. For semantic tokens, we use the textual tokens from Flan-T5~\cite{raffel2020exploring} $\mathbf{T}^\text{gt}_\text{semantic}$ as the supervision and apply MSE loss:
\begin{equation}
    \mathcal{L}_\text{mse}=\Vert \mathbf{T}^\text{gt}_\text{semantic}-\mathbf{T}_\text{semantic}\Vert_2^2.
    \label{eq:t5-mse}
\end{equation}

The residual tokens are employed to supplement the missing information of the semantic tokens. Then, both types of tokens are merged and fed into as the conditional inputs of DiT. Here, we adopt multi-head cross-attention to merge these two tokens and obtain the resultant bridge queries:
\begin{equation}
    {\mathbf{H}}_\text{bridge}=\texttt{Cross-Attn}(\mathbf{T}_\text{semantic},\mathbf{T}_\text{residual},\mathbf{T}_\text{residual}).
    \label{eq:attention}
\end{equation}
For audio generator with ${\mathbf{H}}_\text{bridge}$ as condition, we employ flow-matching~\cite{esser2024scaling} for generative modeling:
\begin{equation}
    \mathcal{L}_\text{flow}=\mathbb{E}_{\boldsymbol{x}_1,\boldsymbol{x}_0,t}\Vert u(\boldsymbol{x}_t,t,\boldsymbol{c})-\boldsymbol{v}_t\Vert_2^2,
    \label{eq:flow-matching}
\end{equation}
where $\boldsymbol{c}$ is the condition and we choose $\boldsymbol{c}={\mathbf{H}}_\text{bridge}$ and $t$ is uniformly sampled from $[0,1]$.
Through the generative supervision, $\mathbf{T}_\text{residual}$ can capture detailed information and complement $\mathbf{T}_\text{semantic}$.

\subsection{Progressive Training Strategy}
\label{subsec:train}

After establishing an effective bridge between the LLM and DiT, it becomes essential to design an efficient end-to-end training mechanism to build synergy between the understanding and generation tasks.
We propose a progressive training strategy, following a single-to-multi and generation-to-unification paradigm. The training could be divided into three stages, where the model (1) learn to generate single audio segments, followed by (2) unified generation and understanding for single audios and (3) long-audio adaptation.

\noindent \textbf{Stage-I: Single-audio generation.}
There are two sub-stages. (1) Stage-I-Warm, AudioStory learns to generate semantic tokens with MSE supervision in Eq.~\eqref{eq:t5-mse}. Only the LoRA of the LLM and the projector of $\mathbf{T}_\text{semantic}$ are updated.
(2) Stage-I-Whole, AudioStory regresses bridge queries based on the input caption, \ie, generating $\mathbf{T}_\text{semantic}$ and $\mathbf{T}_\text{residual}$, respectively. They are subsequently merged via Eq.~\eqref{eq:attention} and fed into DiT. Here, the regression of $\mathbf{T}_\text{semantic}$ and the prediction of its beginning and end tokens are supervised. We tune LoRA of the LLM, all projectors, the attention layer and the generation model DiT. The learning objectives are shown below:
\begin{align}
    \mathcal{L}_{s_1}^\text{warm}=\mathcal{L}_\text{mse},\qquad
    \mathcal{L}_{s_1}^\text{whole}=\mathcal{L}_\text{mse}+\lambda_1\mathcal{L}_\text{text}^\text{token}+\lambda_2\mathcal{L}_\text{flow},
\end{align}
where $\mathcal{L}_\text{text}^\text{token}$ is only applied to the start and the end tokens of $\mathbf{T}_\text{semantic}$.
After this Stage-I, AudioStory possesses a strong capability for single-audio generation.

\noindent \textbf{Stage-II: Single-audio unified generation and understanding.}
Building upon Stage-I, we further introduce audio understanding data to enable unified generation and understanding of single-audio clips. The model takes audio as input for understanding. We freeze the audio encoder while the trainable parameters remain the same as Stage-I-Whole. The learning objectives are in Eq~\eqref{eq:loss-s2}.
\begin{equation}
    \mathcal{L}_{s_2}=\mathcal{L}_\text{mse}+\lambda_1\mathcal{L}_\text{text}+\lambda_2\mathcal{L}_\text{flow}.
    \label{eq:loss-s2}
\end{equation}
With this unified training, AudioStory's generation abilities can be further enhanced.

\noindent \textbf{Stage-III: Long-audio unified generation and understanding.}
We extend the unified training in Stage-II to long-form audio. We further introduce Interleaved Reasoning Generation (Sec.~\ref{subsec:inter-reason-gen}) for narrative audio generation. We curate a high-quality multi-audio dataset to perform supervised fine-tuning. For the generation task, the model sequentially infers the number of audio events based on the input instruction, analyzes the audio content, and performs interleaved generation of captions, semantic tokens, and residual tokens. For the audio continuation task, given the input audio and instruction, the model comprehends the inputs, reasons the key events with story details, and finally generates several short audio segments in a clip-by-clip manner. The audio understanding data incorporates audio Q\&A and instruction data. We keep the learnable components the same as Stage-II. The overall learning objectives are:
\begin{equation}
    \mathcal{L}_{s_3}=\mathcal{L}_\text{mse}+\lambda_1\mathcal{L}_\text{text}+\lambda_2\mathcal{L}_\text{flow}+\lambda_3\mathcal{L}_\text{reason}.
    \label{eq:loss-s3}
\end{equation}

\section{Experiments}
In this section, we first present the experimental setup (Sec.~\ref{subsec:exp-setup}). Then, we compare AudioStory with existing TTA and unified models on long-form audio generation (Sec.~\ref{subsec:exp-long}). We also study the audio understanding and the audio generation (Sec.~\ref{subsec:exp-short}) ability of AudioStory in short audio clips, showing its superior fundamental ability. Finally, in Sec.~\ref{subsec:exp-ablation}, we conduct an in-depth exploration of reasoning forms, bridging query types, joint training strategies, and the synergy between understanding and generation, and provide several key insights.

\subsection{Experimental Setup}
\label{subsec:exp-setup}

\noindent \textbf{Implementation details.}
We choose Qwen-2.5-3B-Instruct~\cite{yang2024qwen2.5} as the LLM and employ DiT pretrained from TangoFlux~\cite{hung2024tangoflux} as the initialization. For encoding instruction for the audio continuation task, we employ Whisper-large-v3~\cite{radford2023robust} as the audio encoder. The projector has two layers with GeLU activations. In Stage-I, AudioStory is trained with lr$=2e^{-4}$ for 50 epochs with a per-device batch size of 32. In Stage-II, we use lr=1e-4 for 10 epochs. The ratio of understanding and generation data is 2:1. In Stage-III, we set different learning rates for LLM and DiT. We set $\lambda_1=1,\lambda_2=0.2,\lambda_3=0.4$. The tunable parameters three-stage training are LoRAs in LLMs, projectors, the cross-attention fuser for bridging queries, and DiT. More details are shown in the Appendix.

\noindent \textbf{Training datasets.}
The training dataset comprises the understanding dataset, single-audio generation and multi-audio (long-audio) generation datasets. For the understanding dataset, we integrated AudioSetCaps~\cite{gemmeke2017audio}, VGGSound~\cite{chen2020vggsound}, MusicCaps~\cite{agostinelli2023musiclm}, and Auto-ACD~\cite{sun2024auto}, converting their captions into QA format. Additionally, we incorporated AudioSetCaps-QA and VGGSound-QA datasets, resulting in 1M audio-QA pairs in total. For the single-audio generation dataset, we combined AudioSetCaps, VGGSound, MusicCaps~\cite{agostinelli2023musiclm}, and Auto-ACD, resulting in 700k audio-caption pairs. For the multi-audio generation dataset, we curated the AudioStory-10k dataset, with details provided in Sec.~\ref{sec:audiostory-10k}.
In Stage-I, we train the model on we train the model on single-audio generation datasets. Stage-II further incorporates the audio understanding dataset beyond Stage-I. As for Stage-III, our model is trained using multi-audio generation as well as understanding datasets.

\noindent \textbf{Evaluation metrics.}
For single-audio generation, we employ the AudioLDM-eval toolkit\footnote{\url{https://github.com/haoheliu/audioldm\_eval}} to compute Frechet Distance (FD), Frechet Audio Distance (FAD)~\cite{kilgour2018fr}, KL-Divergence (KL), and stable-audio-metrics\footnote{\url{https://github.com/Stability-AI/stable-audio-metrics}} for FD$_\text{openl3}$~\cite{cramer2019look}, KL$_\text{passt}$~\cite{koutini2021efficient}, and CLAP score on AudioCaps testset~\cite{kim-etal-2019-audiocaps}. For audio understanding, we consider the tasks of audio question answering (AQA), and audio captioning on AudioCaps and Clotho dataset~\cite{drossos2020clotho}, reporting SPIDEr, CIDEr, and ACC scores. The evaluation metrics for long-audio generation is presented in Sec.~\ref{sec:audiostory-10k}.

\noindent \textbf{Baseline methods.}
Prior audio generation models could be divided into two groups: pure TTA models like AudioLDM2~\cite{liu2024audioldm2learningholistic} and TangoFlux~\cite{hung2024tangoflux} and LLM-based unified models, including CoDi~\cite{tang2023codi} and NExT-GPT~\cite{wu2024nextgpt}. Both of them could only generate short audio clips. For long-form audio generation, we curated three classes of baselines: (1) Directly generating audios with maximum available durations using the whole textual caption. (2) Incorporating LLMs to reason and generate captions for each short audio clip, which are then fed into baseline models to generate multiple audio clips separately. These clips are then concatenated to constitute the final long-form audio. (3) Directly using the ground truth captions in the benchmark, serving as the oracle setting and upper bound for baseline models. In addition, we also report the performance of AudioStory on the Tom \& Jerry and the audio continuation task.

\begin{table}[!t]
\setlength\tabcolsep{3pt}
\centering
\renewcommand{\arraystretch}{1.2}
\caption{Comparative results on long-audio generation. ``Instruct'' is short for instruction-following and ``CLAP'' denotes CLAP score, ``gt'' denotes ground-truth. ``Consis.'' and ``Coher.'' are short for consistency and coherence. Here, \textbf{bold} and \underline{underline} indicate the best and the second-best results.}
\vspace{5pt}
\label{tab:main-multi-audio}
\resizebox{1\textwidth}{!}{
\begin{tabular}{@{}lcccccccc@{}}
\toprule
\multirow{2}{*}{Model} & \multicolumn{3}{c}{Instruction-Following} & \multicolumn{2}{c}{Consistency} & \multicolumn{2}{c}{Generation Quality} & \multicolumn{1}{l}{\multirow{2}{*}{Max. Duration $\uparrow$}} \\ \cmidrule(lr){2-4} \cmidrule(lr){5-6} \cmidrule(lr){7-8}
 & Instruct. $\uparrow$ & CLAP $\uparrow$ & Reasoning $\uparrow$ & Consis. $\uparrow$ & Coher. $\uparrow$ & FD $\downarrow$ & FAD $\downarrow$ & \multicolumn{1}{l}{} \\ \midrule
\textcolor{gray}{AudioLDM2}~\cite{liu2024audioldm2learningholistic} & \textcolor{gray}{2.8} & \textcolor{gray}{0.296} & \textcolor{gray}{-} & \textcolor{gray}{4.6} & \textcolor{gray}{4.4} & \textcolor{gray}{3.43} & \textcolor{gray}{4.49} & \textcolor{gray}{10s} \\
\textcolor{gray}{TangoFlux}~\cite{hung2024tangoflux} & \textcolor{gray}{3.2} & \textcolor{gray}{0.317} & \textcolor{gray}{-} & \textcolor{gray}{4.1} & \textcolor{gray}{4.2} & \textcolor{gray}{2.48} & \textcolor{gray}{3.49} & \textcolor{gray}{30s} \\ \midrule
\textcolor{gray}{Caps (gt)+TangoFlux}~\cite{hung2024tangoflux} & \textcolor{gray}{4.0} & \textcolor{gray}{0.348} & \textcolor{gray}{-} & \textcolor{gray}{2.4} & \textcolor{gray}{2.0} & \textcolor{gray}{1.79} & \textcolor{gray}{3.59} & \textcolor{gray}{30s} \\
LLM+TangoFlux~\cite{hung2024tangoflux} & {\ul 3.5} &  {\ul 0.322} & {\ul 3.5} & 2.1 & 1.9 & {\ul 2.55} & {\ul 3.82} & {\ul 30s} \\
LLM+CoDi~\cite{tang2023codi} & 3.2 & 0.286 & {\ul 3.5} & 1.4 & 1.4 & 3.39 & 4.04 & 10s \\
LLM+NExT-GPT~\cite{wu2024nextgpt} & 3.3 & 0.299 & {\ul 3.5} & 1.8 & 1.7 & 3.47 & 3.99 & 10s \\ \midrule
\RC{30}AudioStory &  \textbf{4.1} & \textbf{0.392} & \textbf{4.2} & \textbf{4.1} & \textbf{3.9} & \textbf{1.43} & \textbf{3.00} & \textbf{150s} \\ \bottomrule
\end{tabular}
}
\end{table}

\begin{table}[!t]
\begin{minipage}[t]{0.49\textwidth}
    \setlength\tabcolsep{3pt}
    \centering
    \renewcommand{\arraystretch}{1.2}
    \caption{Single audio understanding performance.}
    \label{tab:single-understanding}
    \resizebox{\linewidth}{!}{
    \begin{tabular}{@{}lcccccc@{}}
    \toprule
    \multirow{2}{*}{Model} & \multicolumn{2}{c}{ClothoCaps} & \multicolumn{2}{c}{ClothoAQA} & \multicolumn{2}{c}{AudioCaps} \\ \cmidrule(l){2-3} \cmidrule(l){4-5} \cmidrule(l){6-7} 
     & SPIDEr & CIDEr & ACC & B-ACC & SPIDEr & CIDEr \\ \midrule
    UIO-2 XXL~\cite{lu2023unifiedio2} & 5.7 & 6.5 & - & - & - & 48.9 \\
    CoDi~\cite{tang2023codi} & 6.2 & 7.3 & - & - & 48.0 & 78.9 \\
    NExT-GPT~\cite{wu2024nextgpt} & 13.8 & 20.3 & 26.4 & 39.5 & 53.4 & 80.7 \\
    Spider~\cite{lai2024spider} & - & - & - & - & 53.7 & 81.9 \\ \midrule
    \RC{30}AudioStory-Base & \textbf{24.1} & \textbf{37.7} & \textbf{42.8} & \textbf{60.6} & \textbf{54.8} & \textbf{83.2} \\ \bottomrule
    \end{tabular}
    }
\end{minipage}
\begin{minipage}[t]{0.49\textwidth}
    \setlength\tabcolsep{3pt}
    \centering
    \renewcommand{\arraystretch}{1.05}
    \caption{Single audio generation performance.}
    \label{tab:single-generation}
    \resizebox{\linewidth}{!}{
    \begin{tabular}{@{}lcccccc@{}}
    \toprule
    \multirow{2}{*}{Model} & \multicolumn{6}{c}{AudioCaps Test Set} \\
    \cmidrule(l){2-7} 
    & FD$_\text{openl3}$ $\downarrow$ & KL$_\text{passt}$ $\downarrow$ & FD $\downarrow$ & FAD $\downarrow$ & KL $\downarrow$ & CLAP $\uparrow$ \\ \midrule
    Make-An-Audio~\cite{huang2023makeanaudiotexttoaudiogenerationpromptenhanced} & 128.49 & 1.16 & 1.65 & 3.16 & 0.63 & 0.256 \\
    stable-audio-open~\cite{evans2024stableaudioopen} & 103.68 & 1.12 & 1.63 & 2.98 & 0.61 & 0.298 \\
    AudioLDM2~\cite{liu2024audioldm2learningholistic} & 87.74 & 1.01 & {\ul 1.59} & 2.63 & 0.57 & 0.252 \\
    TangoFlux~\cite{hung2024tangoflux} & {\ul 83.58} & {\ul 0.95} & 1.57 & {\ul 2.34} & {\ul 0.52} & \textbf{0.385} \\
    CoDi~\cite{tang2023codi} & 121.66 & 1.17 & 1.69 & 9.61 & 0.60 & 0.228 \\
    NExT-GPT~\cite{wu2024nextgpt} & 107.18 & 1.13 & 1.64 & 5.69 & 0.59 & 0.265 \\ \midrule
    \RC{30}AudioStory-Base & \textbf{83.39} & \textbf{0.91} & \textbf{1.52} & \textbf{2.29} & \textbf{0.51} & {\ul 0.383} \\ \bottomrule
    \end{tabular}
    }
\end{minipage}
\vspace{-10pt}
\end{table}

\subsection{Long-Form Narrative Audio Generation}
\label{subsec:exp-long}

\noindent \textbf{Instruction-following ability.}
As shown in Table~\ref{tab:main-multi-audio}, considering the instruction-following aspect, AudioStory demonstrates a significant advantage in complex scenarios involving multiple events and sounding objectives. It outperforms the LLM-aided TTA models by 17.85\% on the CLAP score, thereby demonstrating the superior instruction-following generation capability of our model. Our method effectively addresses the issue of overlooking sounding entities, which can be attributed to the enhanced understanding and decomposition of the instruction. 

\noindent \textbf{Generation quality.} AudioStory demonstrates strong long-form audio generation performance across both natural scenes and the cartoon domain. Our approach achieves superior FD and FAD scores compared to diffuser-based and LLM+diffuser baselines. This improvement is reasonable: (1) we enhance long-form audio generation by single-audio clip training, effectively extending high-quality short-audio generation to longer sequences; (2) the duration of generated audio is longer and more closely matches the reference long audios than those generated by previous methods.

\noindent \textbf{Consistency.}
\emph{Notably, consistency is meaningful only with strong instruction-following.} For example, AudioLDM2, despite high consistency scores from short (10s) outputs, performs poorly on instruction-following, making it a weak baseline. In contrast, our method achieves substantial advantages in both consistency and coherence, reaching scores of 4.0 and 3.7, respectively, as in Table~\ref{tab:main-multi-audio}. It is worth noting that in the consistency evaluation, AudioStory achieves comparable performance despite generating significantly longer audio with richer narratives compared to TTA models.

\subsection{Single-Audio Generation}
\label{subsec:exp-short}

\noindent \textbf{Joint audio generation \& understanding.}
We also evaluate our model’s performance on short audio generation and understanding tasks, and conduct comparisons with TTA and LLM-based models. For the generation task in Table~\ref{tab:single-generation}, AudioStory outperforms prior competitors on both suites of evaluation tools, even outperforming the state-of-the-art TTA model, \ie, TangoFlux~\cite{hung2024tangoflux}, indicating the effectiveness of the proposed LLM and DiT bridging mechanism. As for the audio understanding task in Table~\ref{tab:single-understanding}, AudioStory outperforms advanced LLM-based models, which means that our method could competently handle both generation and understanding tasks.

\subsection{Qualitative Analysis}

AudioStory exhibits strong reasoning capabilities, it can accurately divide the input instruction into several events based on narrative logic and temporal order, subsequently generating short audio clips segment by segment, and ultimately composing a coherent long-form audio. Besides, AudioStory could accurately infer the duration of each audio clip. Here, we provide a qualitative case in Fig.~\ref{fig:case-duration1}. More cases are presented in the Appendix~\ref{sec:appendix-qualitative}.

\begin{figure}[!t]
    \centering
    \includegraphics[width=1.0\linewidth]{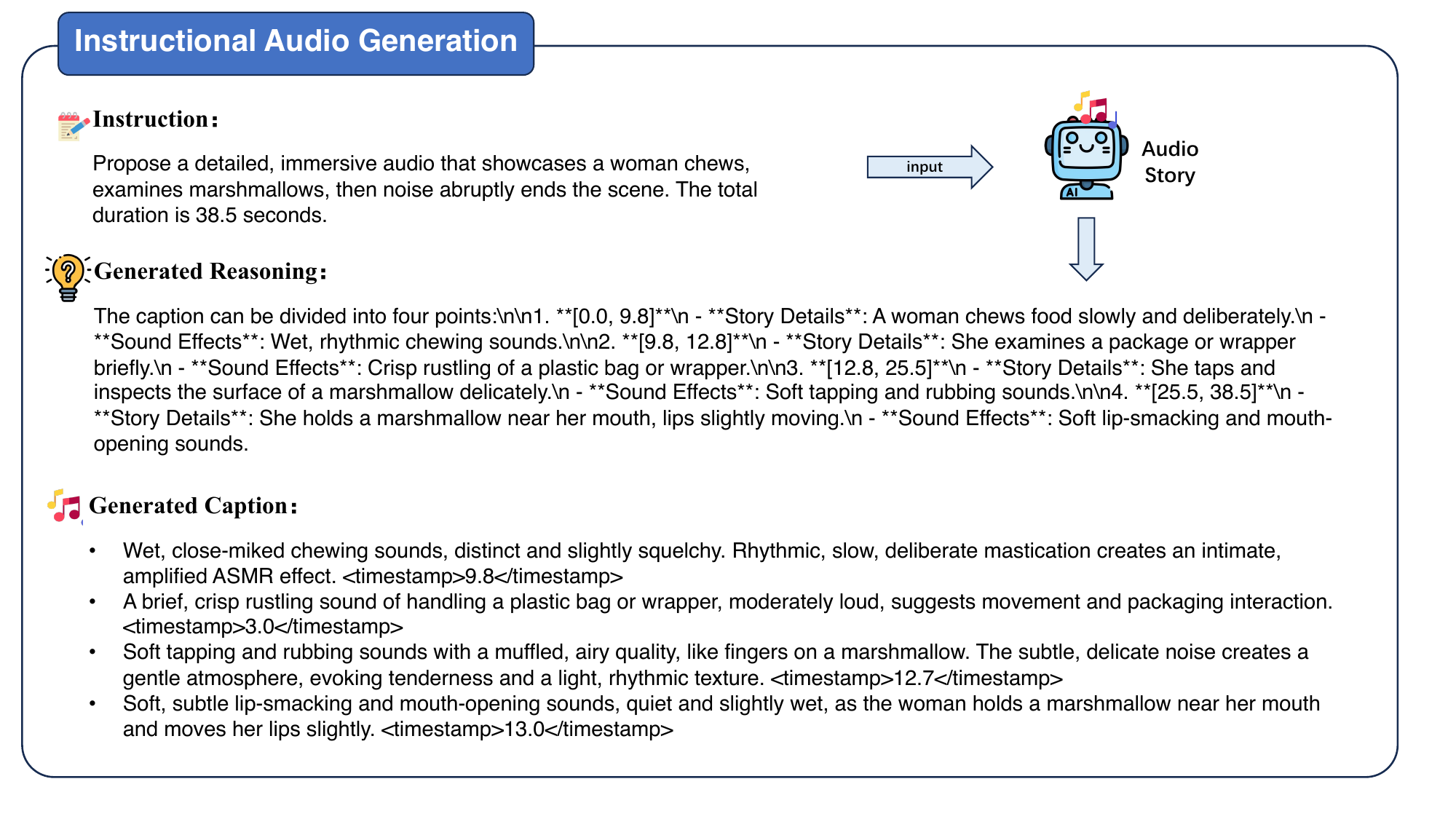}
    \vspace{-10pt}
    \caption{Qualitative case of long-form audio generation.}
    \label{fig:case-duration1}
\end{figure}

\subsection{Ablation Studies}
\label{subsec:exp-ablation}

\begin{wraptable}{R}{0.4\textwidth}
\setlength\tabcolsep{2pt}
\centering
\renewcommand{\arraystretch}{0.7}
\scriptsize
\caption{Ablations of reasoning.}
\vspace{-7pt}
\label{tab:ablation-reasoning}
\resizebox{\linewidth}{!}{
\begin{tabular}{@{}ccccc@{}}
\toprule
Variant & Cons. $\uparrow$ & Inst. $\uparrow$ & FAD $\downarrow$ & CLAP $\uparrow$ \\ \midrule
w/o reasoning & 3.1 & 3.1 & 4.13 & 0.34 \\
w/o interleaved & 1.6 & 1.2 & 16.03 & 0.14 \\
w/ reasoning & \textbf{4.0} & \textbf{4.1} & \textbf{3.06} & \textbf{0.39} \\ \bottomrule
\end{tabular}
}
\end{wraptable}
\paragraph{Does interleaved reasoning generation help narrative audio generation?}

We progressively investigate which forms of reasoning are effective for long-form narrative audio generation. We ablate two model variants: (a) a model that skips instruction analysis and directly generates captions for audio clips, and (b) a model that performs instruction decomposition without explicitly generating audio captions. As shown in Table~\ref{tab:ablation-reasoning}, ablating without reasoning leads to simplified audio events with missing objects or actions, resulting in a significant drop in both instruction-following performance and CLAP score. When interleaved reasoning is removed, the model can still infer event content, but lacks contextual guidance when generating bridge queries, severely degrading audio quality. We conclude that reasoning is essential for narrative audio generation, with interleaved reasoning being the most critical component. Explicit caption generation for each audio clip is necessary to ensure generation quality.

\begin{wraptable}{R}{0.45\textwidth}
\setlength\tabcolsep{3pt}
\centering
\renewcommand{\arraystretch}{0.7}
\scriptsize
\caption{Ablations on bridging mechanism.}
\vspace{-7pt}
\label{tab:ablation-bridge}
\resizebox{\linewidth}{!}{
\begin{tabular}{@{}ccccc@{}}
\toprule
ID & BQ & Sup. Feat. & Single & Multi \\ \midrule
(a) & \multirow{2}{*}{Semantic} & AudioMAE~\cite{huang2022masked} & 9.55 & 11.39 \\
(b) &  & Whipser~\cite{radford2023robust} & 10.26 & 12.31 \\ \midrule
(c) & \multirow{2}{*}{Residual} & AudioMAE~\cite{huang2022masked} & 9.24 & 10.06 \\
(d) &  & Whisper~\cite{radford2023robust} & 11.06 & 11.21 \\ \midrule
(e) & \multirow{2}{*}{\begin{tabular}[c]{@{}c@{}}Residual\\ +guid.\end{tabular}} & AudioMAE~\cite{huang2022masked} & 3.60 & 4.21 \\
(f) &  & Whisper~\cite{radford2023robust} & 3.71 & 4.39 \\ \midrule
\RC{30}(g) & Ours & T5 w/o guid. & \textbf{2.29} & \textbf{3.12} \\ \bottomrule
\end{tabular}
}
\end{wraptable}

\noindent \textbf{Which type of features are suitable for bridging between the LLM and the DiT?}
Our analysis suggests that audio features, on one hand, have lower semantic density, and on the other hand, are more difficult for the LLM to fit compared to textual features, especially given the complex temporal structure in Whisper. Therefore, supervising the semantic tokens with textual features is more suitable and efficient. For residual tokens, Table~\ref{tab:ablation-bridge} (c)–(g) shows that explicit or weak supervision using existing audio features significantly harms generation performance.
In summary, text features with rich semantics are well-suited for supervising semantic tokens, while for residual tokens, applying weak supervision through the DiT loss is the most effective way to capture complementary low-level audio information.

\noindent \textbf{What are the key factors in end-to-end joint training of unified models?}
For unified models, prior arts typically train the LLM and DiT separately, connecting them via a zero-shot bridging mechanism, which results in a feature gap. To address this, we propose end-to-end joint training of the LLM and DiT.
We begin by focusing on the end-to-end training paradigm, as in Table~\ref{tab:ablation-dit}. Notably, when residual tokens are removed, overall performance drops significantly. Analysis reveals that the LLM and DiT focus on different types of information, and directly updating the LLM using the DiT loss severely impairs its performance. In contrast, residual tokens effectively alleviate this issue.
Secondly, we explore how to configure DiT’s learnable parameters. As shown in (c)–(f), fully freezing DiT degrades performance, while fully unfreezing it achieves the best results. Notably, unfreezing MM-DiT outperforms Single-DiT, since the latter focuses on low-level features that are more sensitive to noise, thus affecting generation quality.
Thus, we can draw the following conclusions:
(1) End-to-end joint training of the LLM and DiT is essential.
(2) Residual tokens play a critical role, as they capture low-level complementary information and help mitigate conflicts between DiT and LLM during optimization.
(3) Fully unfreezing DiT is necessary. Selectively unfreezing either the Single-DiT or MM-DiT \emph{alone} leads to suboptimal performance.


\begin{table}[!t]
\setlength\tabcolsep{3pt}
\centering
\renewcommand{\arraystretch}{1}
\caption{Ablations on the end-to-end joint training strategy of DiT. Here ``S-DiT'' and ``M-DiT'' denote Single-DiT and MM-DiT. ``Consis.'' denotes consistency.}
\label{tab:ablation-dit}
\resizebox{.85\textwidth}{!}{
\begin{tabular}{@{}cccccccccc@{}}
\toprule
\multirow{2}{*}{ID} & \multirow{2}{*}{\begin{tabular}[c]{@{}c@{}}Semantic\\ Tokens\end{tabular}} & \multirow{2}{*}{\begin{tabular}[c]{@{}c@{}}Residual\\ Tokens\end{tabular}} & \multirow{2}{*}{\begin{tabular}[c]{@{}c@{}}DiT Joint\\ Training\end{tabular}} & \multirow{2}{*}{\begin{tabular}[c]{@{}c@{}}Tunable\\ Module\end{tabular}} & 
\multicolumn{3}{c}{Single Audio} & \multicolumn{2}{c}{Multi Audio} \\ \cmidrule(l){6-8} \cmidrule(l){9-10} 
 &  &  &  &  & FD $\downarrow$ & FAD $\downarrow$ & KL $\downarrow$ & Consis. $\uparrow$ & FAD $\downarrow$ \\ \midrule
(a) & \cmark & \xmark & \xmark & - & 1.57 & 2.33 & 0.52 & 3.2 & 5.23 \\
(b) & \cmark & \xmark & \cmark & open all & 2.16 & 4.66 & 0.84 & 3.4 & 4.98 \\
(c) & \cmark & \xmark & \cmark & freeze & 4.86 & 11.04 & 0.89 & 1.3 & 12.97 \\
(d) & \cmark & \cmark & \cmark & open S-DiT & {2.37} & {5.84} & {0.64} & {2.1} & {6.28} \\
(e) & \cmark & \cmark & \cmark & open M-DiT & {1.98} & {3.21} & {0.67} & {3.5} & {3.64} \\ 
(f) & \cmark & \cmark & \cmark & open all & \textbf{1.53} & \textbf{2.29} & \textbf{0.51} & \textbf{4.3} & \textbf{3.00} \\ \bottomrule
\end{tabular}
}
\end{table}
\noindent \textbf{How to progressively build the synergy between generation and understanding?}
Both understanding and generation training are essential to our model, making a progressive training strategy crucial. Here, we examine the effectiveness of various progressive training approaches. 
In Table~\ref{tab:ablation-train}, without progressive training, performance on both comprehension and generation drops significantly, even worse than training the tasks independently, which is primarily due to the inherent conflict and task interference between them. In contrast, a well-structured progressive strategy enables unified training to outperform isolated approaches, highlighting its necessity.
Further exploration into synergizing generation and comprehension reveals key insights: training generation first, then adding comprehension, yields optimal overall performance, with strong comprehension accuracy. In contrast, reversing the order harms generation, and interleaved training similarly undermines overall optimization.
Therefore, we conclude that generation and understanding exhibit inherent synergy, and their optimal training order depends on the primary objective. For unified generation-understanding models, training generation first and then introducing understanding is the most effective strategy. In contrast, omitting progressive training or interleaving both tasks impairs overall optimization.

\begin{table}[!t]
\setlength\tabcolsep{3pt}
\centering
\renewcommand{\arraystretch}{.9}
\caption{Ablations on progressive training. ``Gen.'', ``Und.'' and ``BQ'' denote generation, understanding and Bridge Queries. ``SAG'' and ``LAG'' are short for single and long-form audio generation.}
\label{tab:ablation-train}
\resizebox{.8\textwidth}{!}{
\begin{tabular}{@{}ccccccccc@{}}
\toprule
\multirow{2}{*}{ID} & \multirow{2}{*}{Order} & \multirow{2}{*}{Stage-I} & \multirow{2}{*}{Stage-II} & \multirow{2}{*}{Stage-III} & SAG & LAG & \multicolumn{2}{c}{Audio Und.} \\ \cmidrule(l){6-7} \cmidrule(l){8-9} 
 &  &  &  &  & FAD $\downarrow$ & FAD $\downarrow$ & CIDEr $\uparrow$ & SPIDEr $\uparrow$ \\ \midrule
(a) & \multirow{3}{*}{Und.$\to$Gen.} & Und. & - & - & - & - & 35.7 & 23.1 \\
(b) &  & Und. & BQ & - & 7.42 & 9.53 & 36.9 & 23.8 \\
(c) &  & Und. & BQ & DiT joint & 6.50 & 7.26 & \textbf{38.6} & \textbf{24.9} \\ \midrule
(d) & \multirow{4}{*}{Gen.$\to$Und.} & BQ & - & - & 2.37 & 5.23 & - & - \\
(e) &  & BQ & Und. & - & {\ul 2.35} & {\ul 4.98} & 31.5 & 19.5 \\
(f) &  & BQ & Und. & DiT joint & 3.61 & 6.50 & 24.6 & 16.4 \\
(g) &  & BQ & DiT joint & Und. & \textbf{2.29} & \textbf{3.00} & {\ul 37.7} & {\ul 24.1} \\ \midrule
(h) & N/A & \multicolumn{3}{c}{DiT joint + Und.} & 5.70 & 8.74 & 27.3 & 18.2 \\ \bottomrule
\end{tabular}
}
\end{table}

\subsection{Human Evaluation}

\noindent \textbf{Evaluation protocol.} Beyond API-based evaluation, we further conducted an anonymous user study on our model and baseline models. We employ 30 participants to manually score a total of 150 audio clips, generated from 50 instructions, by our model, Tangoflux, and Next-GPT, respectively. The participants listened to the long-form audio generated by different models based on the same instruction. They scored the audio on four criteria: instruction-following, consistency, generation quality, and reasoning logic. The scores were averaged to compute user consistency.
As shown in the Table~\ref{tab:human_eval}, AudioStory consistently outperforms other competitors in terms of instruction-following, consistency, quality and reasoning logic.

\paragraph{Correlation between Gemini-based \& human-based evaluation.}

Qualitatively, human evaluation results show our model performs the best among all three models, with the LLM + TTA model outperforming the LLM + any-to-any model. This aligns with the results from our Gemini evaluation.
Quantitatively, we analyze the correlation between the human subjective and Gemini-based objective evaluation. We calculate Cohen's kappa coefficient between these two evaluation protocols. Specifically, we compute the correlation across two dimensions, \ie, different comparative methods in Table~\ref{tab:human_eval} and different test samples. The results in Table~\ref{tab:cohen-kappa} indicate a high correlation between the human and Gemini scoring distributions across various models and samples, validating the correctness of the proposed Gemini-based evaluation.

\begin{table}[!t]
\centering
\small

\begin{minipage}[t]{0.7\textwidth}
\setlength\tabcolsep{3pt}
\renewcommand{\arraystretch}{1}
\centering
\caption{Human evaluation of the generated audios for different methods on: instruct-following, consistency, fidelity, and reasoning logic.}
\vspace{2pt}
\renewcommand\arraystretch{1.0}
\setlength{\tabcolsep}{1.5 mm}{
\begin{tabular}{@{}lcccc@{}}
\toprule
Method            & Instruct-Follow & Consist. & Fidelity & Reason. Logic \\ \midrule
LLM + TangoFlux   & 3.52               & 3.22        & 3.58               & 3.19            \\
LLM + NExT-GPT    & 3.10               & 2.56        & 2.87               & 3.14            \\
\RC{30}AudioStory (Ours) & 4.23               & 4.68        & 4.37               & 4.22            \\ \bottomrule
\end{tabular}
}

\label{tab:human_eval}

\end{minipage}
\hfill
\begin{minipage}[t]{0.28\textwidth}
\setlength\tabcolsep{3pt}
\small

\centering
\caption{Correlation of Gemini and human scores.}
\vspace{2pt}
\renewcommand\arraystretch{1.85}
\setlength{\tabcolsep}{1.5 mm}{
\begin{tabular}{@{}ccc@{}}
\toprule
 & \makecell{Across\\model} & \makecell{Across\\model} \\
 \midrule
Kappa Coef. & 0.91         & 0.83         \\ \bottomrule 
\end{tabular}
}

\label{tab:cohen-kappa}
\end{minipage}
\end{table}
\section{Extended Applications}

\begin{figure}[!t]
    \centering
    \includegraphics[width=.95\linewidth]{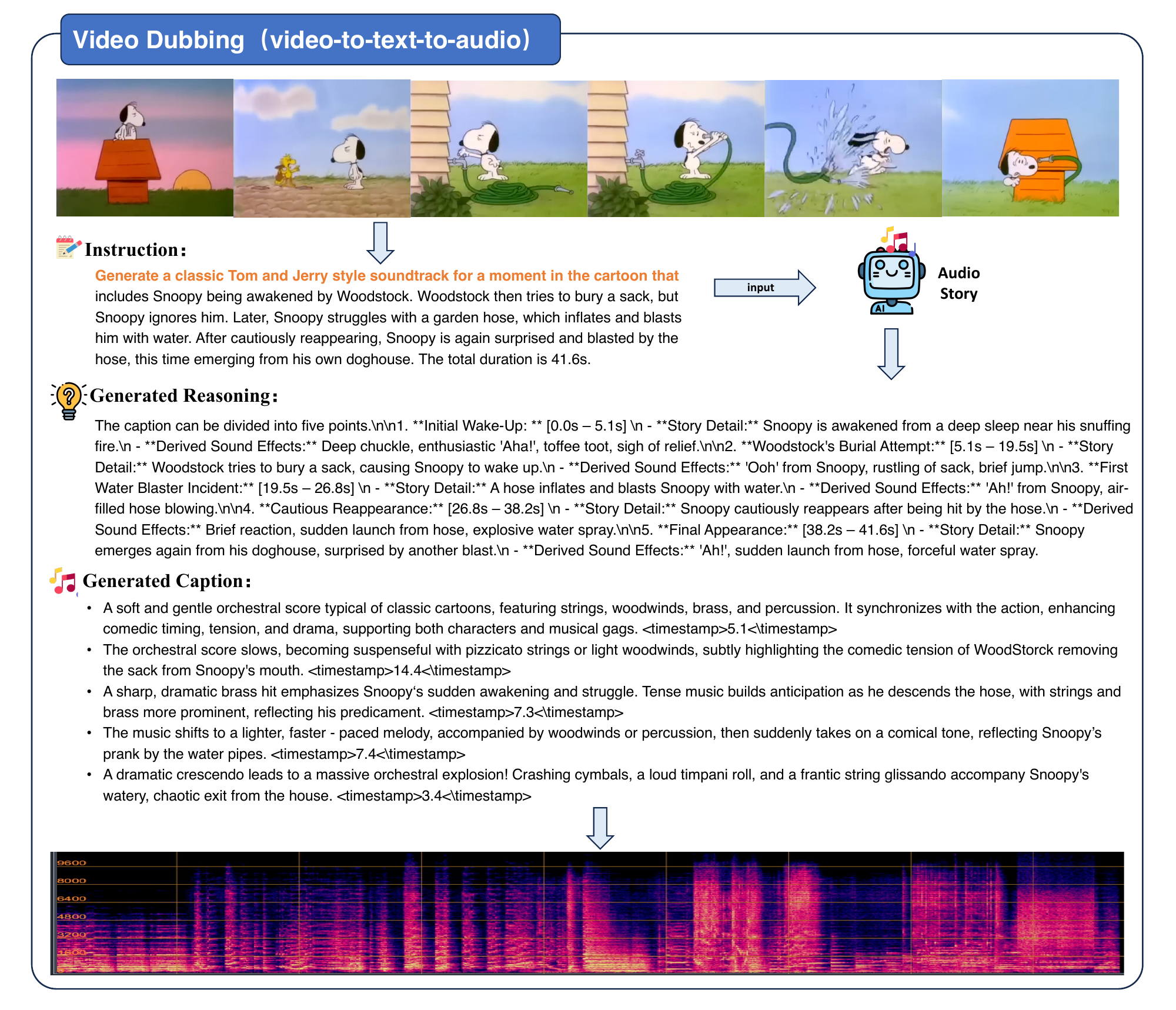}
    \vspace{3pt}
    \caption{Case of naive video dubbing: First, we extract captions from the video, then write the extracted captions as instructions and send them to AudioStory for audio generation.}
    \label{fig:video-dubbing-text}
    \vspace{-5pt}
\end{figure}

\begin{figure}[!t]
    \centering
    \includegraphics[width=.95\linewidth]{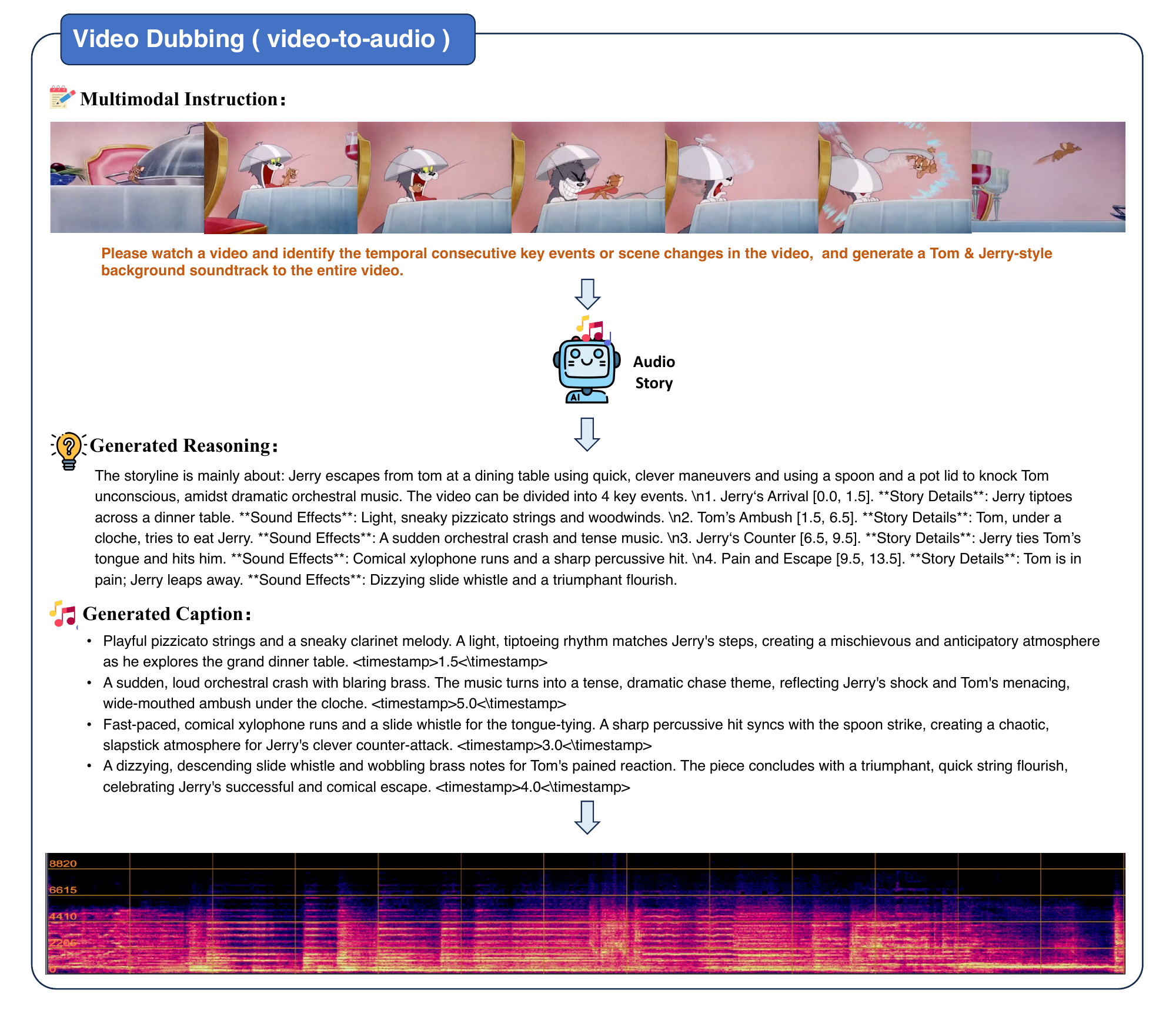}
    \vspace{3pt}
    \caption{Case of video dubbing: We input both the video and the instruction into the model, which parses the narrative into segments, extracts story details with corresponding audio elements, and sequentially generates aligned audio clips..}
    \label{fig:video-dubbing-video}
    \vspace{7pt}
\end{figure}

\begin{figure}[!t]
    \centering
    \includegraphics[width=.95\linewidth]{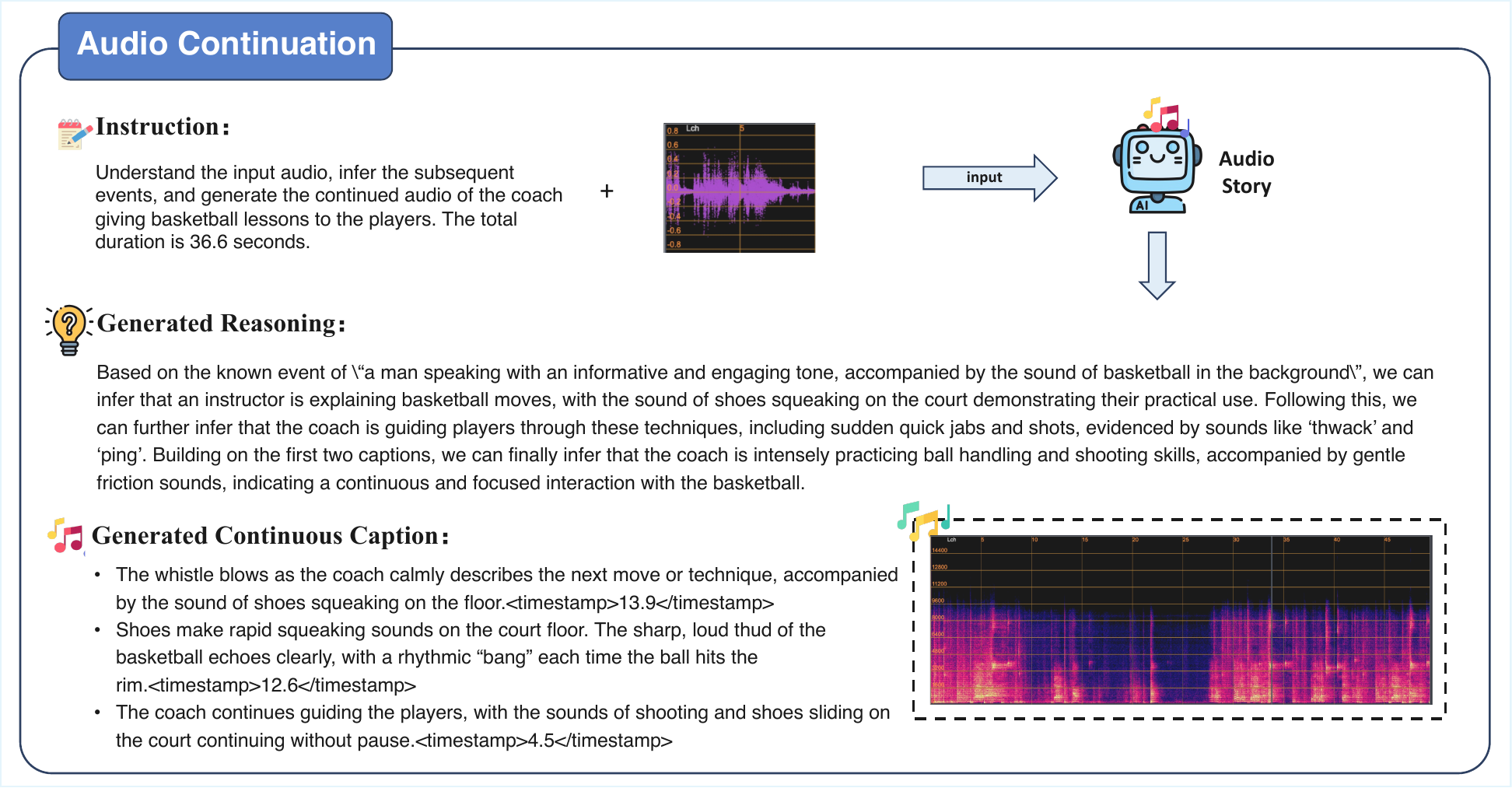}
    \vspace{7pt}
    \caption{Qualitative cases of audio continuation.}
    \label{fig:cases-3}
    \vspace{7pt}
\end{figure}

\begin{figure}[!t]
    \centering
    \includegraphics[width=.95\linewidth]{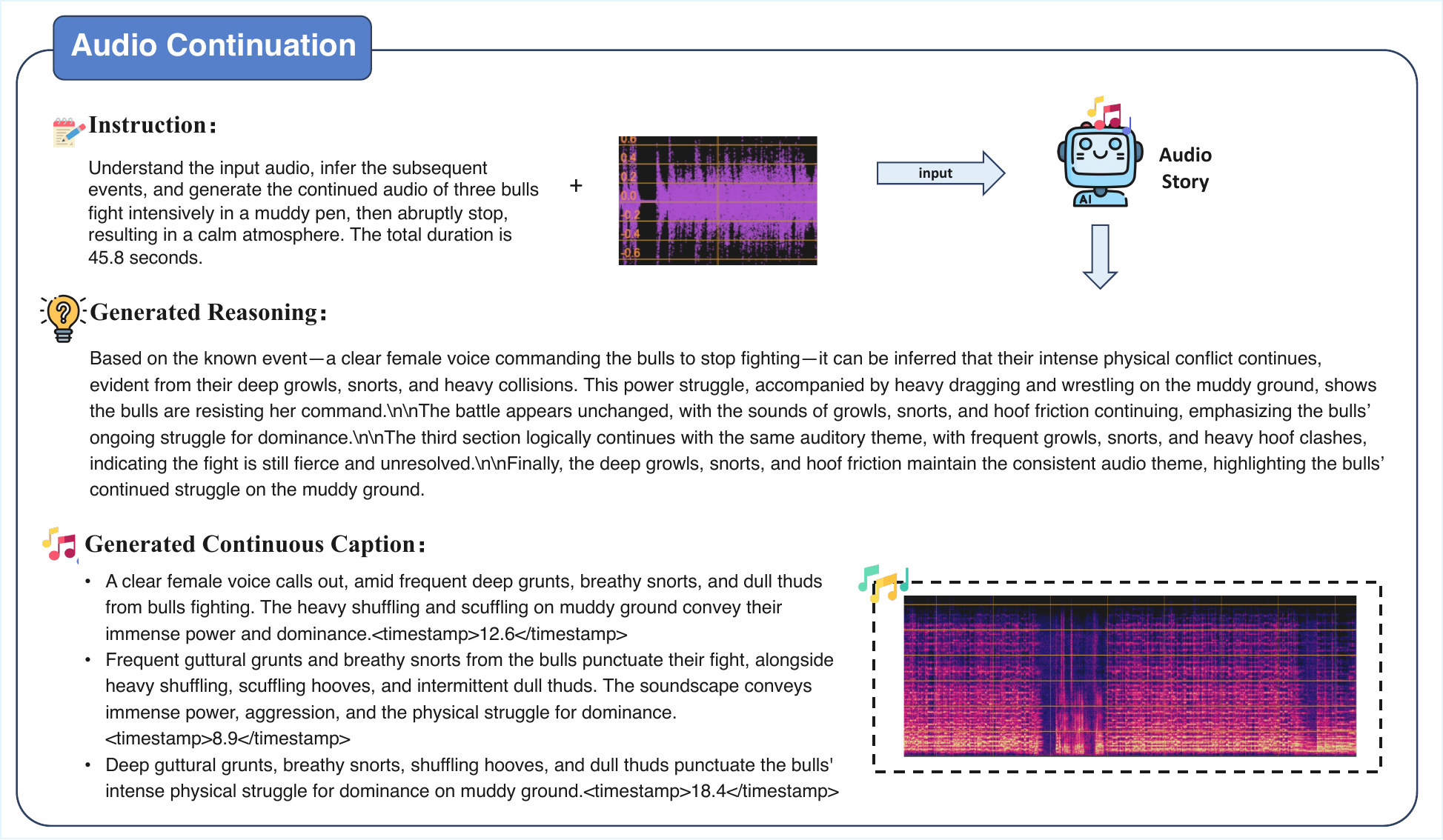}
    \caption{Qualitative cases of audio continuation.}
    \label{fig:cases-4}
\end{figure}

We present AudioStory’s extended applications, including instruction-conditioned narrative audio generation, audio continuation, and video dubbing. Specifically, AudioStory generates long-form narrative audio from diverse multimodal instructions, remains closely aligned with input video and directives, and exhibits robust instruction-following.

\paragraph{Video dubbing.} While prior experiments focused on text-based instructions, we extend AudioStory to a more practical scenario: video dubbing. This enhancement lets the model thoroughly analyze video content, reason about event sequences with timestamps, and generate synchronized audio. An initial approach uses Gemini-2.5-pro to generate a video summary caption, then performs instruction-based long-form narrative audio generation for the given video, as shown in Fig.~\ref{fig:video-dubbing-text}.

However, this method poorly aligns audio with visual content. Therefore, we further modified the model to accept both video data and instructions as input. Specifically, the LLM reasons about the video to generate bridging tokens: it first comprehends the video's overall content, then sequentially decomposes it into temporally ordered events, inferring each event's visual details and corresponding audio information. Technically, we replace the LLM with a pretrained video MLLM (\ie, Qwen2.5-VL~\cite{bai2025qwen2.5vl}) and jointly train the LLM and audio generator using LoRA tuning. The training data is from the animated sound partition of AudioStory-10k. We provide the video dubbing results in Fig.~\ref{fig:video-dubbing-video}.

\vspace{-7pt}
\paragraph{Audio continuation.}
Given an audio segment and an instruction, our model performs audio continuation. AudioStory first reasons about the content of the subsequent audio to be generated, then proceeds with segment-by-segment generation. The cases are shown in Fig.~\ref{fig:cases-3} and Fig.~\ref{fig:cases-4}.

\section{Conclusion}

In this paper, we tackle the key limitations of existing text-to-audio and unified models in generating long-form narrative audio in complex scenarios. We introduce AudioStory, a unified understanding-generation model endowed with robust multimodal instruction-following and reasoning capabilities. To achieve this, we design an interleaved reasoning generation process, a decoupled bridging mechanism, and a progressive training strategy that jointly leverage the reasoning power of LLMs and strengthen the synergy between understanding and generation. Additionally, we present AudioStory-10k, the first benchmark for long-form narrative audio generation, which includes fine-grained annotations of audio and audio-visual events with timestamps and detailed reasoning trajectories. Our comprehensive analyses cover reasoning forms, bridge query types, end-to-end training strategies for LLM-DiT integration, and the collaborative dynamics between understanding and generation, providing practical insights for future model development.

\vspace{-7pt}
\paragraph{Limitations and Future Work.}
Since multimodal instruction for long audio generation remains underexplored, future work can incorporate more sophisticated designs, \eg, integrating multiple audio generators to better address the issue of overlapping audio segments.
We also consider blending text and audio generation within the same autoregressive multimodal LLM, as well as delve into the relationship and synergy between audio generation and understanding.

{
\small
\bibliography{ref}
\bibliographystyle{unsrt}
}

\clearpage

\title{AudioStory: Generating Long-Form Narrative Audio with Large Language Models}
\appendix

\maketitle

\section*{\large Appendix}

\section{Implementation Details}

We provide detailed hyper-parameters of three training stages in Table~\ref{tab:appendix-params}. In Stage-II and Stage-III, the ratio of generation and understanding samples is 2:1. For LLM, we choose Qwen2.5-VL-3B-Instruct and only tune LoRA to avoid overfitting. TangoFlux is employed as the initialization of DiT for audio generation. For the weights of different loss functions, we set the weight of $\mathcal{L}_\text{mse}$ for T5 regression, $\mathcal{L}_\text{text}$ for next-token-prediction and $\mathcal{L}_\text{flow}$ for DiT as 5, 2 and 1, respectively.

\begin{table}[!h]
\setlength\tabcolsep{3pt}
\centering
\renewcommand{\arraystretch}{1.8}
\caption{Detailed hyper-parameters of three training stages. Here, ``A'' denotes audio, ``proj.'' and ``lr'' are short for the projector and learning rate. We use 16 GPUs and report the overall batch size.}
\label{tab:appendix-params}
\resizebox{\textwidth}{!}{
\begin{tabular}{cc|cc|c|c}
\hline
\multicolumn{2}{c|}{\multirow{2}{*}{Dimension}} & \multicolumn{2}{c|}{Stage-I} & \multirow{2}{*}{Stage-II} & \multirow{2}{*}{Stage-III} \\ \cline{3-4}
\multicolumn{2}{c|}{} & \multicolumn{1}{c|}{Warm-up} & Whole &  &  \\ \hline
\multicolumn{2}{c|}{Task} & \multicolumn{1}{c|}{A$\to$T5} & {A$\to$T5 with DiT}. & {A$\to$T5 with DiT} + Und. & {A$\to$T5 with DiT} + Und. + Reasoning \\ \hline
\multicolumn{2}{c|}{Dataset} & \multicolumn{2}{c|}{AudioCaps, WavCaps} & \begin{tabular}[c]{@{}c@{}}I+AudioSetCaps (Q\&A), VGGSound (Q\&A),\\ MusicCaps, Auto-ACD\end{tabular} & AudioStory-10k \\ \hline
\multicolumn{1}{c|}{\multirow{2}{*}{Model}} & Trainable & \multicolumn{1}{c|}{LLM, proj. ($\mathbf{T}_\text{semantic}$)} & LLM, all proj., DiT & LLM, all projectors, DiT & LLM, all proj., DiT \\ \cline{2-6} 
\multicolumn{1}{c|}{} & Frozen & \multicolumn{1}{c|}{Whipser, DiT} & Whisper & Whisper & Whisper \\ \hline
\multicolumn{1}{c|}{\multirow{3}{*}{\begin{tabular}[c]{@{}c@{}}Training\\ Config\end{tabular}}} & batch size & \multicolumn{1}{c|}{512} & 256 & Gen.: 8, Und.: 16 & Gen.: 8, Und.: 16 \\ \cline{2-6} 
\multicolumn{1}{c|}{} & lr & \multicolumn{2}{c|}{1e-3} & 1e-4 & LLM (2e-5), DiT (5e-5) \\ \cline{2-6} 
\multicolumn{1}{c|}{} & epoch & \multicolumn{1}{c|}{25} & 25 & 10 & 10 \\ \hline
\end{tabular}
}
\end{table}

\begin{figure}[!h]
    \centering
    \includegraphics[width=1.0\linewidth]{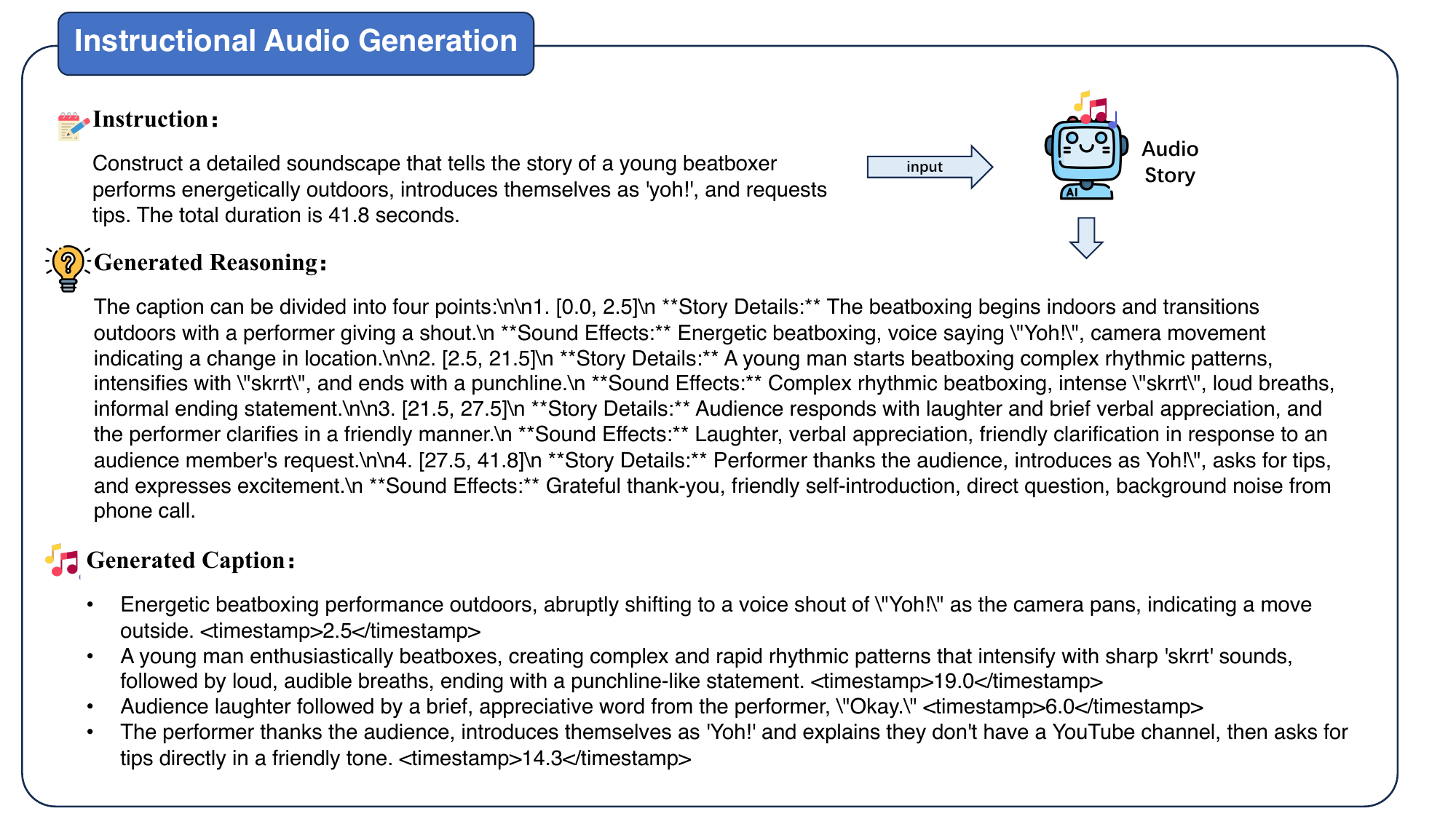}
    \caption{Long-form audio generation case \#2.}
    \label{fig:case-duration2}
\end{figure}

\section{More Qualitative Cases}
\label{sec:appendix-qualitative}

we present more cases for long-form audio generation. Our model could automatically derive the duration of each audio segment to be generated, as shown in Fig.~\ref{fig:case-duration2}. It can be observed that AudioStory could accurately determine the number of events based on the instruction and provide precise descriptions for each audio segment, including both the events themselves and their associated sound effects. Finally, AudioStory can precisely reason out the caption for each segment and generate the corresponding audio clips accordingly.




\begin{figure}[!h]
    \centering
    \includegraphics[width=1.0\linewidth]{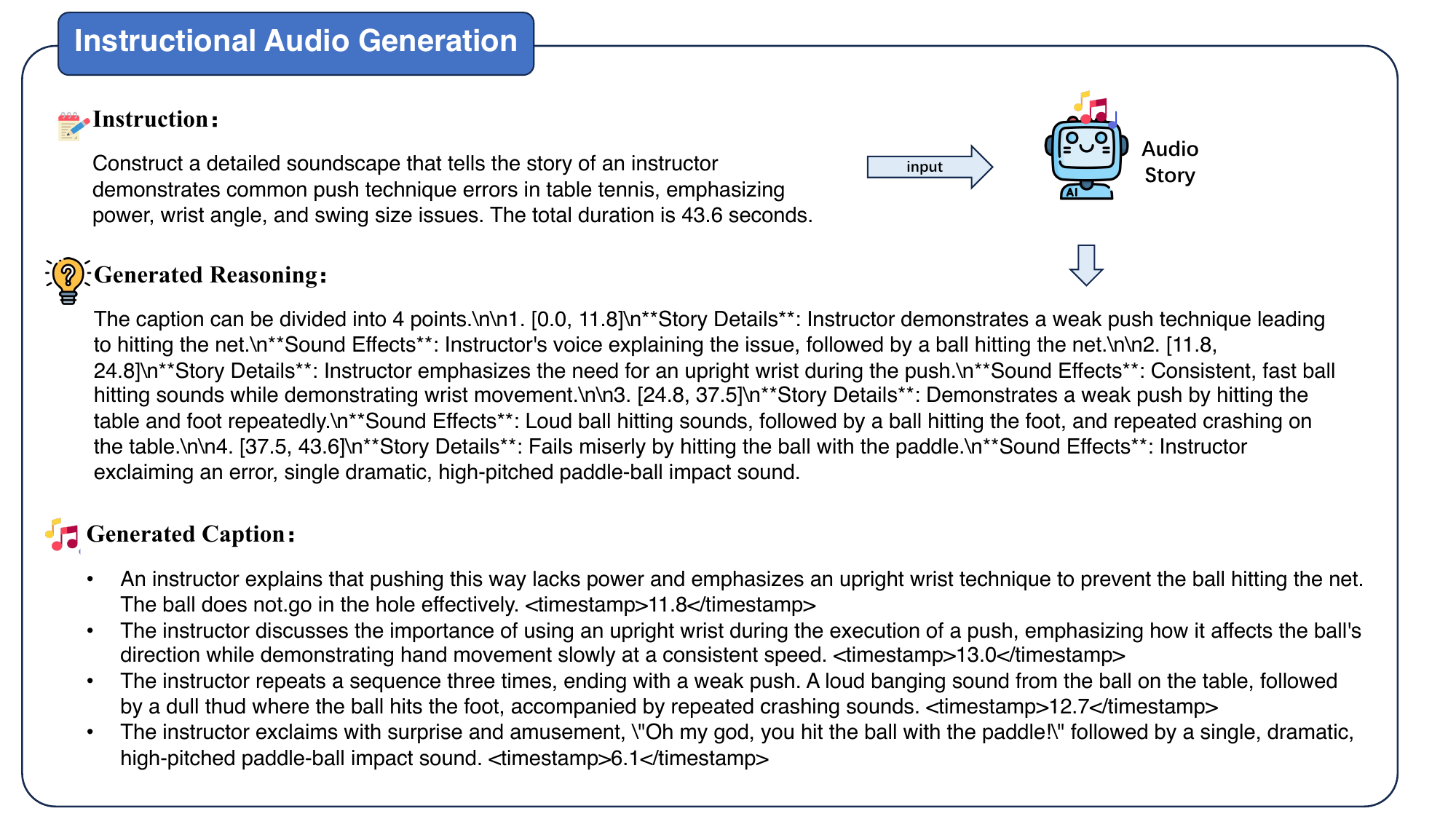}
    \caption{Long-form audio generation case \#3.}
    \label{fig:case-duration3}
\end{figure}

\begin{figure}[!h]
    \centering
    \includegraphics[width=1.0\linewidth]{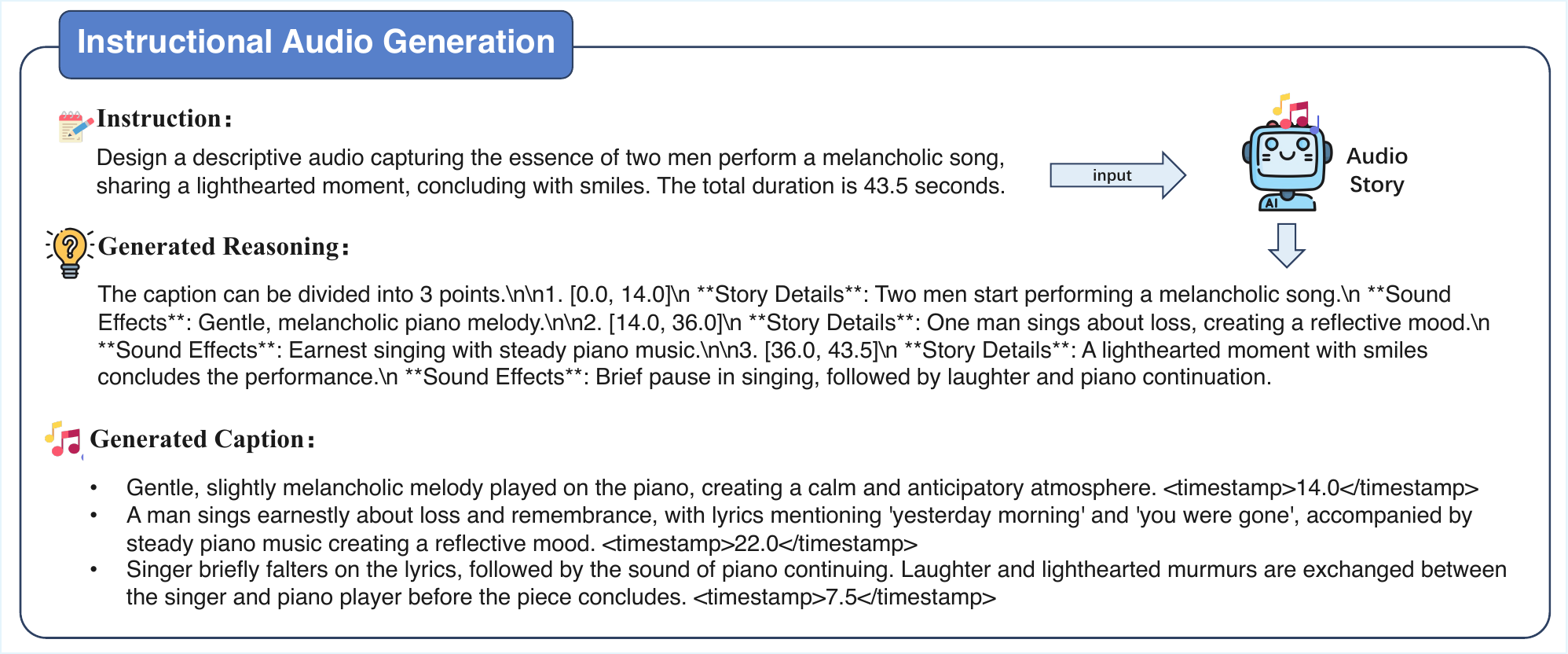}
    \caption{Long-form audio generation case \#3.}
    \label{fig:case-duration5}
\end{figure}


\section{More Explorations of Residual Tokens}
For residual tokens, we not only explore their forms and training strategies, but also investigate hyperparameters such as their quantity and fusion methods with semantic tokens.

\paragraph{The number of residual tokens.}
Here, we study the impact of different numbers of residual tokens, and report both single- and long-form audio generation, as in Table~\ref{tab:appendix-num-tokens}. For single-audio generation, too few residual tokens lead to degraded performance. We attribute this to two factors: less low-level complementary information is captured. Additionally, residual tokens help mitigate conflicts between the LLM and the DiT, while too few tokens weaken this effect. Conversely, an excessive number of tokens also degrades performance, because they increase the difficulty for the LLM to regress. Similar patterns could also be observed in the long-form scenario. Overall, 8 residual tokens are most suitable for both single and long audio scenarios.

\begin{table}[!h]
\setlength\tabcolsep{15pt}
\centering
\renewcommand{\arraystretch}{1}
\caption{Detailed ablations of the number of residual tokens }
\label{tab:appendix-num-tokens}
\resizebox{.7\textwidth}{!}{
\begin{tabular}{@{}ccccc@{}}
\toprule
\multirow{2}{*}{\# Tokens} & \multicolumn{3}{c}{Single Audio} & Long Audio \\ \cmidrule(l){2-4} \cmidrule(l){5-5} 
 & FD $\downarrow$ & FAD $\downarrow$ & KL $\downarrow$ & Consistency $\uparrow$ \\ \midrule
1 & 4.01 & 5.02 & 0.93 & 3.2 \\
4 & 3.64 & 3.95 & 0.96 & 3.9 \\
8 & \textbf{1.53} & \textbf{2.29} & \textbf{0.51} & \textbf{4.1} \\
16 & 3.51 & 3.75 & 0.94 & 3.9 \\ \bottomrule
\end{tabular}
}
\end{table}

\begin{wrapfigure}{R}{0.4\textwidth}
    \centering
    \includegraphics[width=\linewidth]{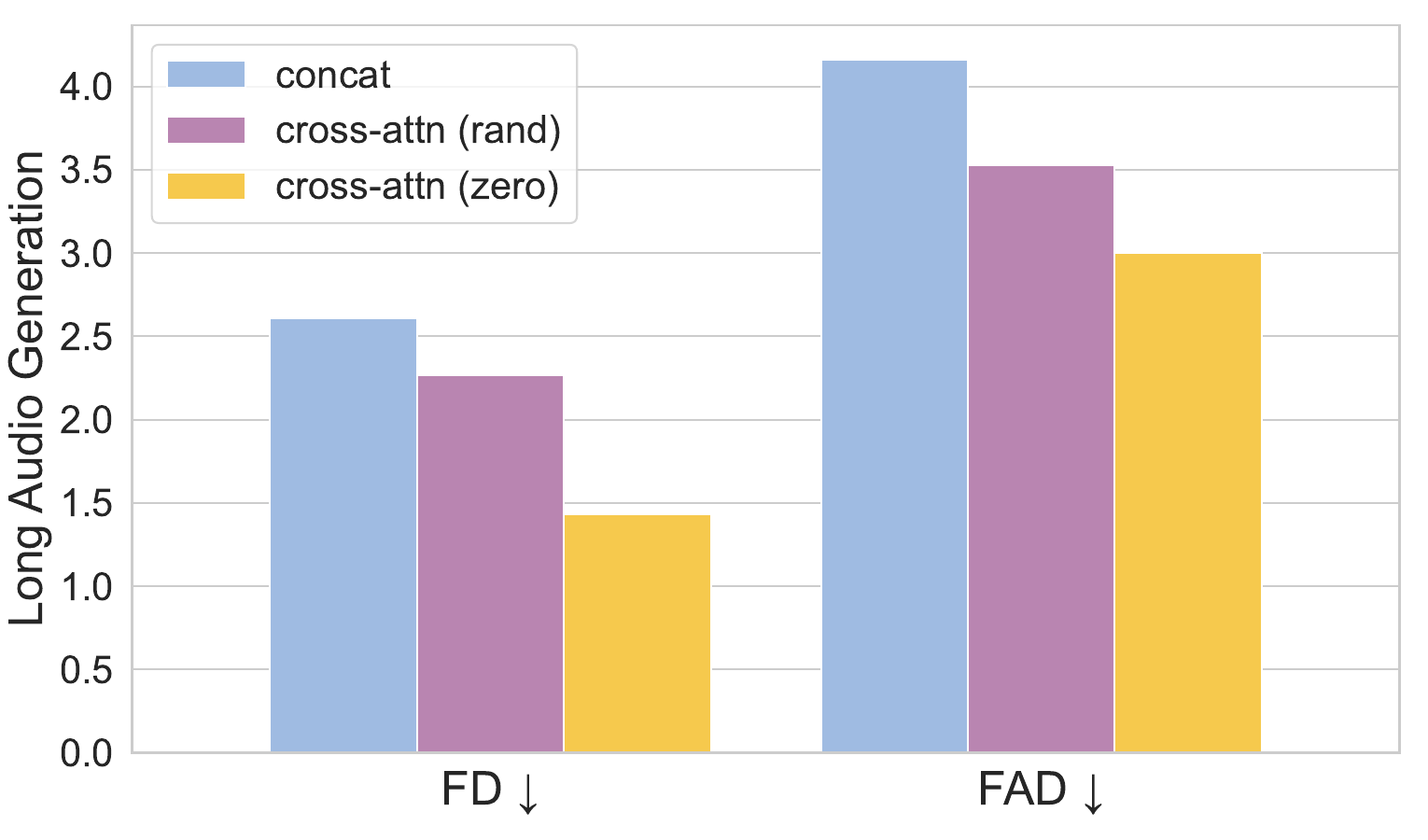}
    \caption{Ablations of token merging.}
    \label{fig:ablation-merging}
\end{wrapfigure}

\paragraph{Merging mechanism of residual tokens.}
For the merging mechanism between semantic and residual tokens, we also conduct in-depth explorations. Here, we mainly consider concatenation and cross-attention. The results of long-form audio generation are reported in Fig.~\ref{fig:ablation-merging}. From the results, one can observe that compared to concatenation, cross-attention ensures more effective fusion of the two features. Additionally, zero-initializing the final layer of the cross-attention module is necessary to prevent excessive disturbance to the semantic tokens at the beginning of training.

\section{What do Residual Tokens Learn?}

\begin{figure}[!t]
    \centering
    \includegraphics[width=1.0\linewidth]{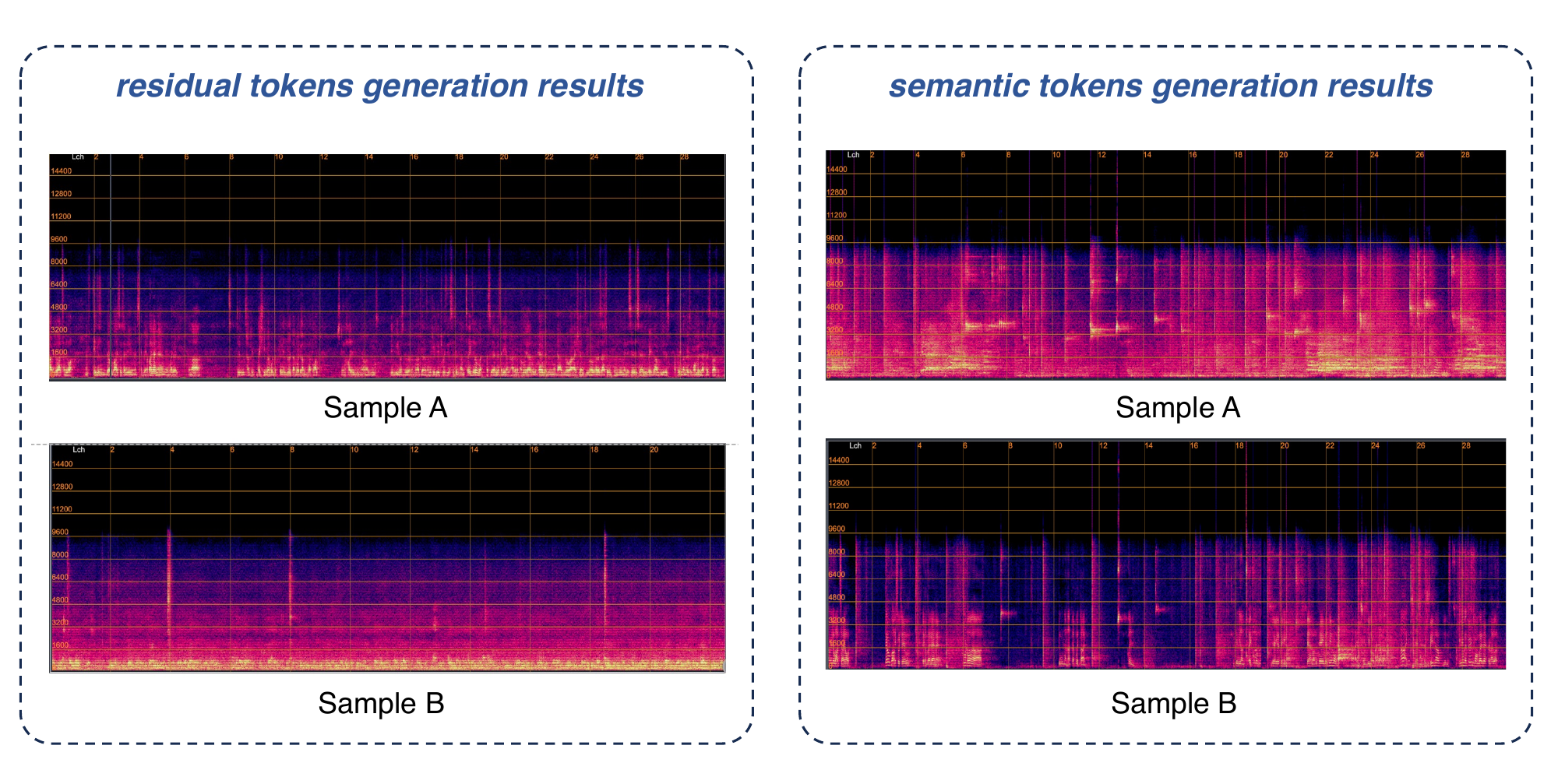}
    \caption{Visualizations of residual tokens.}
    \label{fig:residual}
\end{figure}

To thoroughly explore the effect of residual tokens, we provide visualizations in Fig.~\ref{fig:residual} (left). Specifically, the DiT takes \emph{only} the residual tokens as the input and generates its corresponding audio. We subsequently concatenate all audio clips to constitute the whole long-form audio. The results reveal that for the same audio sample, the residual tokens capture temporally consistent low-level information, primarily reflecting coherence across different audio clips. In contrast, for different samples, the learned residual characteristics vary distinctly. By contrast, semantic tokens learn the underlying global semantics of the input audio and represent the progression of events over time, as illustrated in Fig.~\ref{fig:residual} (right).

\section{AudioStory-10k Benchmark}
\label{appendix:benchmark}

\subsection{Dataset construction pipeline}
\label{appendix:benchmark-dataset}
\begin{figure}[!t]
    \centering
    \includegraphics[width=1.0\linewidth]{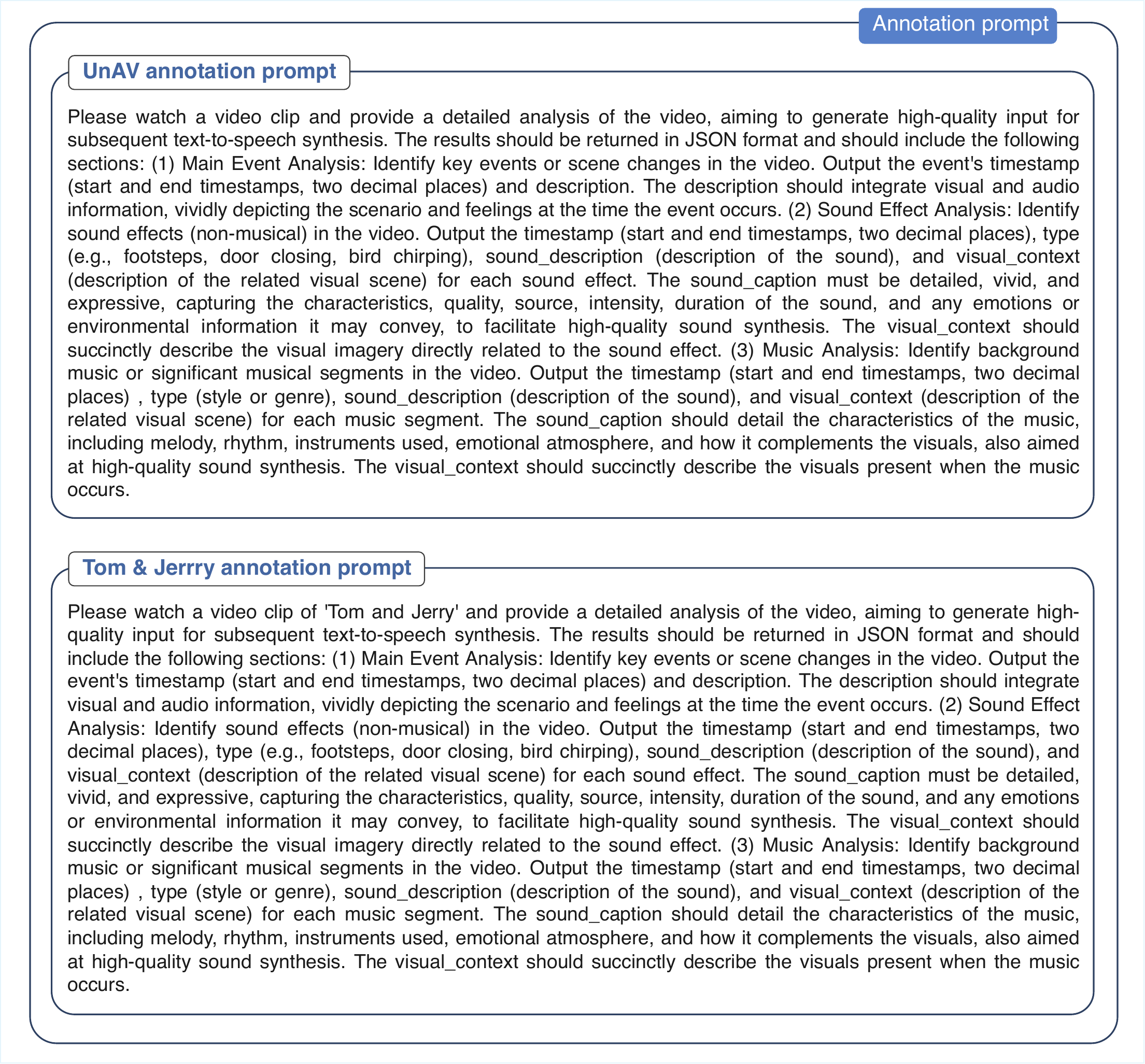}
    \caption{AudioStory-10k annotation prompts.}
    \label{fig:datasets_annotate_prompt}
\end{figure}

The dataset construction pipeline is illustrated as follows. First, we filter videos to select those containing continuous audio events with visually grounded storylines. Next, in the event parsing stage, we use Gemini-2.0-flash to decompose each video into multiple key audio events, each annotated with a timestamp, audio caption, and visual caption, as in Fig.~\ref{fig:datasets_annotate_prompt}. Finally, we perform instruction generation: based on fine-grained textual annotations, GPT-4o is used to generate diverse narrative instructions, accompanied by reasoning steps including task decomposition, audio event timeline planning, scene transitions, and emotional tone inference.

\subsection{Benchmark Construction}
\label{appendix:benchmark-eval}
\begin{figure}[!t]
    \centering
    \includegraphics[width=1.0\linewidth]{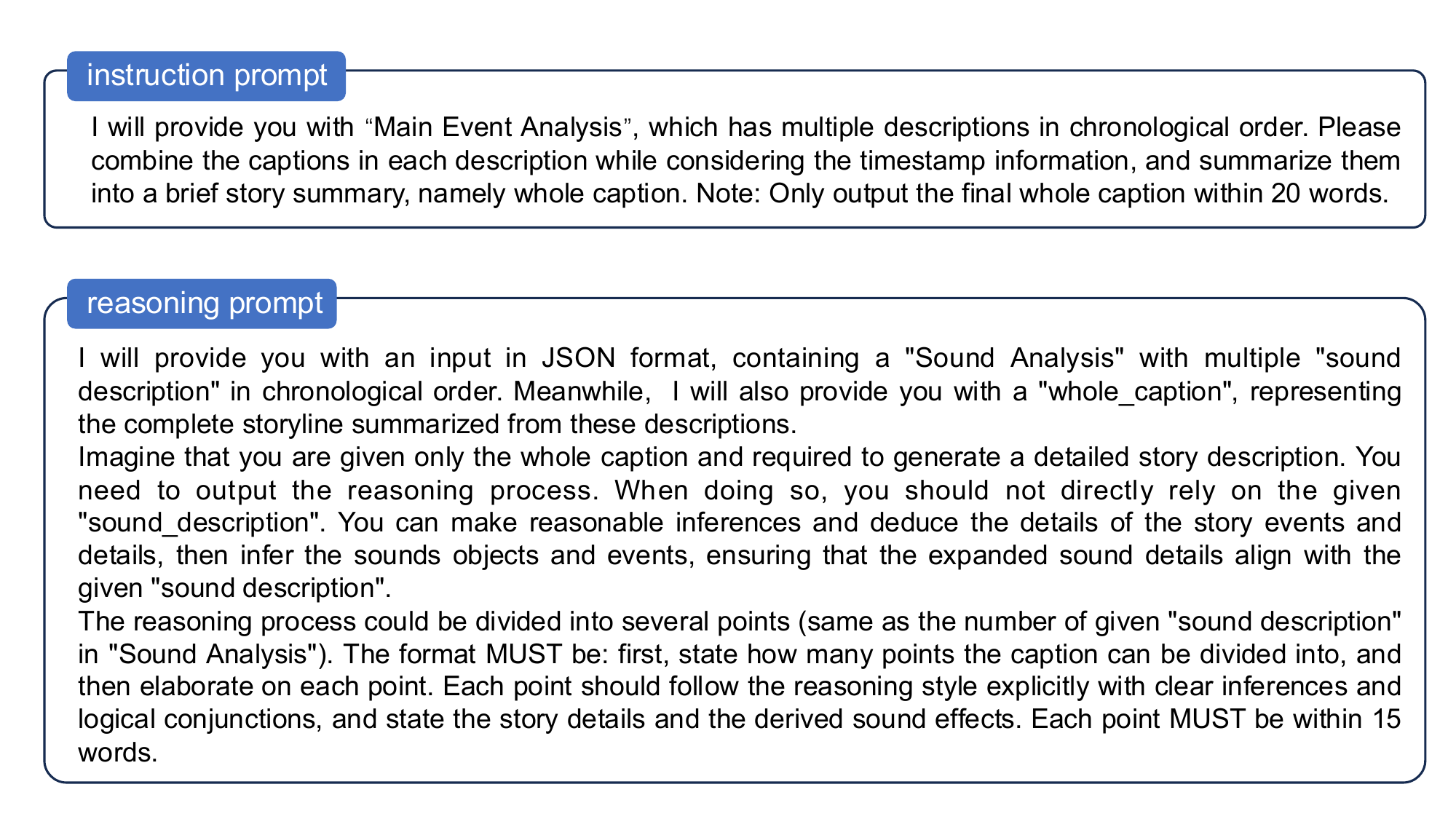}
    \caption{The datasets construction prompt.}
    \label{fig:datasets_prompt}
\end{figure}

\begin{figure}[!t]
    \centering
    \includegraphics[width=1.0\linewidth]{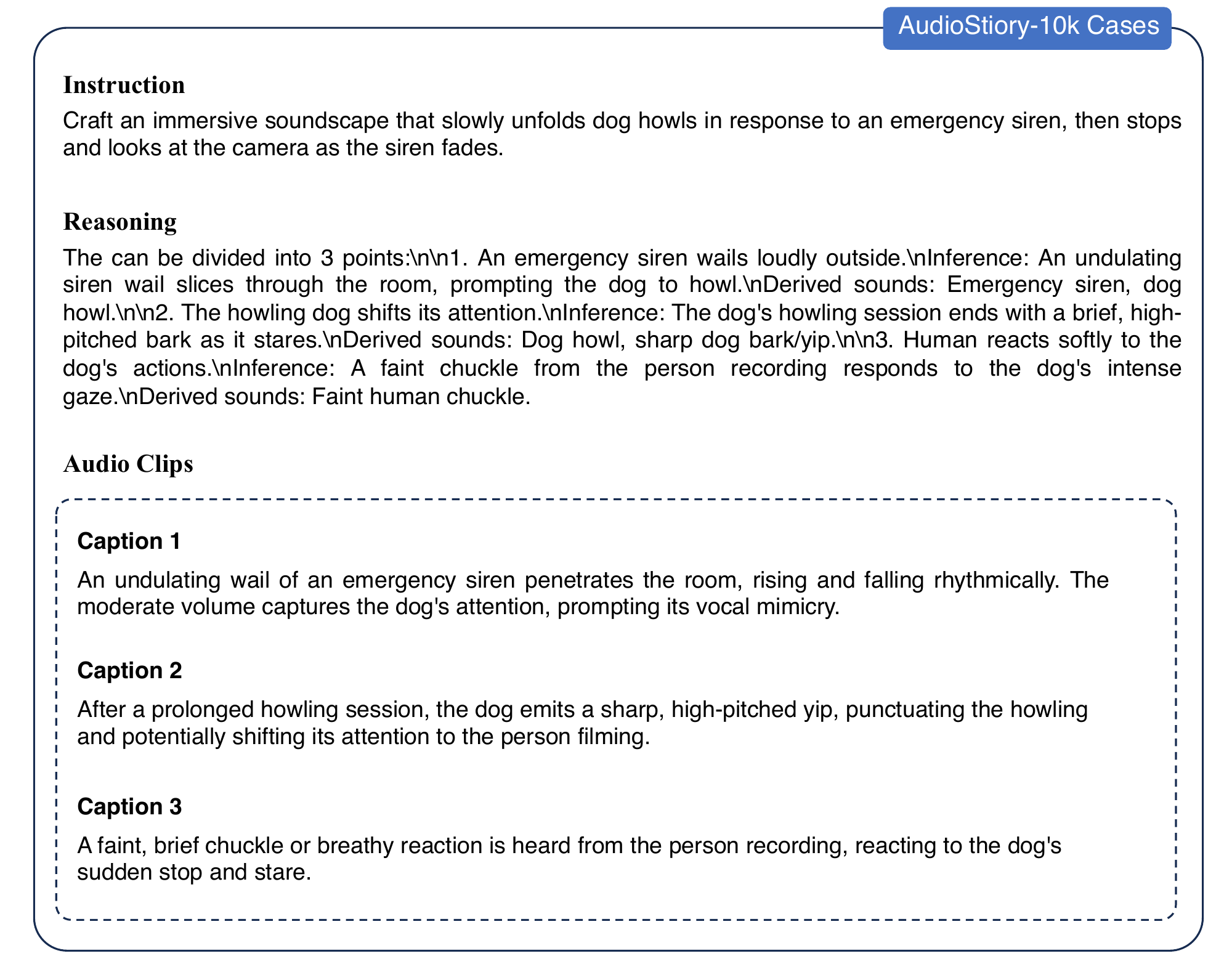}
    \caption{AudioStory-10k cases.}
    \label{fig:datasets_cases}
\end{figure}

\paragraph{Dataset prompt.}
The constructed dataset consists of instructions, reasoning, and audio clips, each with its caption and duration. Specifically, after parsing videos into key audio events using Gemini-2.0-flash as described in Sec.~\ref{sec:audiostory-10k}, we obtain annotations for each event including timestamps, audio captions, visual captions, and audiovisual event captions.
For instruction generation, we use audio-visual event captions as the source input. A prompt, shown in Fig.~\ref{fig:datasets_prompt}, is used to summarize the whole caption of the full audio, which is then incorporated into a predefined instruction template to produce the final instruction.
For reasoning generation, we provide GPT-4o with the whole caption along with the individual captions for each audio clip. GPT-4o is then prompted to infer the reasoning structure. The reasoning consists of two levels: a high-level decomposition indicating how the whole caption can be divided into several parts, followed by detailed descriptions for each part, including the corresponding events and sound-producing content.
An example is illustrated in Fig.~\ref{fig:datasets_cases}.

\paragraph{Benchmark evaluation.}

Along with the curated dataset, we also construct the long-form narrative audio generation task and its associated benchmark. 

(1) Evaluation with Gemini-2.0-flash API, assessing consistency, coherence, instruction following, and reasoning logic.
(2) Evaluation with traditional metrics to measure audio generation quality, including FD, FAD, and CLAP score, among others.

For the Gemini-based evaluation, we design tailored scoring criteria for each metric:

\textbf{(a) Consistency.}
\begin{itemize}
    \item \textbf{Timbre and Sonic Cohesion} Evaluate whether the primary sound sources maintain a generally consistent timbre and unified sonic characteristics.
    \item \textbf{Sound-Producing Entity Consistency} Assess whether the implied sound-producing entities remain consistent, or if changes feel natural and logical within the audio.
    \item \textbf{Acoustic Environment Consistency} Evaluate the background ambience, reverberation, and spatial impression for overall consistency or reasonable progression.
    \item \textbf{Transition Smoothness} Assess whether the transitions between segments are smooth and free of jarring disruptions..
\end{itemize}

\textbf{(b) Coherence.}
\begin{itemize}
    \item \textbf{Intentional Transitions} Check whether transitions between segments are smooth, purposeful, and naturally connected.
    \item \textbf{Dynamic and Emotional Flow} Assess if the dynamic and emotional progression feels consistent or evolves logically, without unjustified sudden shifts.
    \item \textbf{Tempo and Textural Compatibility} Evaluate whether tempo, rhythm, and sonic textures between segments are compatible and blend cohesively.
    \item \textbf{Transition Smoothness} Judge if segment connections are fluid, without abrupt or disjointed
\end{itemize}

\textbf{(c) Instruction following.}
\begin{itemize}
    \item \textbf{Overall Semantic Alignment} Evaluate whether the generated audio broadly reflects the intended scene, actions, and atmosphere described in the instruction. Minor differences are acceptable if the main idea remains clear.
    \item \textbf{Key Element Presence} Verify whether the important sound-producing entities, actions, and environmental elements mentioned in the instruction are reasonably represented. Missing a few non-central elements is acceptable if key parts are present. Additional sounds not specified in the instruction are acceptable if they logically fit the scene and do not disrupt coherence.
    \item \textbf{Event Sequence and Logical Development} Assess whether the overall event progression is reasonable according to the instruction. Small deviations in order are acceptable if they do not break the logical flow.
    \item \textbf{Specific Sound Detail Accuracy} Evaluate whether important sound features (such as types of sounds, tonal qualities, or intensities) are reasonably reflected. Natural variations are acceptable as long as they do not change the overall character of the audio.
\end{itemize}

\textbf{(d) Reasoning logic.}
\begin{itemize}
    \item \textbf{Overall Reasoning Logic} Evaluate whether the model demonstrates a coherent, logical process in interpreting the instruction and planning the audio scene.
    \item \textbf{Caption-Instruction Alignment} Assess whether the generated audio caption accurately reflects the instruction’s key content, sound-producing elements, and described environment.
    \item \textbf{Event Coverage Completeness} Determine whether the inferred and described audio events fully cover the instruction’s core elements, with no major omissions.
    \item \textbf{Semantic and Temporal Accuracy} Evaluate whether the implied timeline and semantic structure of the generated audio align with the instruction’s flow and intent.
\end{itemize}

\subsection{Single-Audio Evaluation Details}
To evaluate the audio generation model, four key metrics assess different aspects of performance:
\begin{itemize}
    \item Frechet Distance (FD) measures the statistical similarity between log-Mel spectrogram distributions of generated and real audio, quantifying low-level spectral fidelity (\eg, pitch, timbre) through mean and covariance comparisons in the mel-spectral domain.
    \item Frechet Audio Distance (FAD) extends FD using high-level embeddings from a pre-trained audio encoder (\eg, VGGish), evaluating perceptual and semantic realism by comparing abstract features like instrument timbre, musical structure, and environmental acoustics.
    \item CLAP Score calculates the cosine similarity between audio and text embeddings from a cross-modal model, assessing how well generated audio aligns with semantic prompts (\eg, textual descriptions of sound content or context).
    \item KL-Divergence (KL) measures the distributional dissimilarity between generated and real audio features (spectral, latent, \etc.), identifying consistency in probability distributions and helping debug issues like mode collapse or over-dispersion in outputs. Collectively, these metrics ensure a comprehensive evaluation of spectral realism, perceptual quality, semantic accuracy, and distributional consistency in generated audio.
\end{itemize}

\end{document}